\documentclass[journal]{IEEEtran}

\usepackage{xcolor,soul,framed} 

\colorlet{shadecolor}{yellow}
\usepackage[pdftex]{graphicx}
\graphicspath{{../pdf/}{../jpeg/}}
\DeclareGraphicsExtensions{.pdf,.jpeg,.png}

\usepackage[cmex10]{amsmath}
\usepackage{array}
\usepackage{mdwmath}
\usepackage{multirow}
\usepackage{booktabs, siunitx}
\usepackage{mdwtab}
\usepackage{eqparbox}
\usepackage{url}
\usepackage{amsfonts}

\usepackage{amssymb}
\usepackage{pifont}
\newcommand{\cmark}{\ding{51}}%
\newcommand{\xmark}{\ding{55}}%

\usepackage{algorithm}
\usepackage{algpseudocode}
\usepackage{subcaption}
\usepackage{longtable}
\usepackage{multicol} 
\usepackage{xcolor}
\usepackage{tablefootnote}
\usepackage{todonotes}
\usepackage{float}
\usepackage[font=small]{caption}
\usepackage{tabularx} 
\usepackage{hyperref}
\usepackage{dirtytalk}
\usepackage{morewrites}
\begin{document}
    \title{Enhancing Imbalanced Node Classification via Curriculum-Guided Feature Learning and Three-Stage Attention Network}
    
  \author{Abdul Joseph Fofanah, ~\IEEEmembership{Student Member,~IEEE,}
        Lian Wen,~\IEEEmembership{Member,~IEEE,}
        David Chen,~\IEEEmembership{Member,~IEEE,}
        Shaoyang Zhang,~\IEEEmembership{Member,~IEEE}

 \thanks{This work was supported in part by Griffith University under Grant 58455.}
  \thanks{Abdul Joseph Fofanah, David Chen, and Lian Wen are with the School of Information and Communication Technology, Griffith University, Brisbane, 4111, Australia. (e-mail:abdul.fofanah@griffithuni.edu.au, orcid=0000-0001-8742-9325;  email: l.wen@griffith.edu.au, orcid=0000-0002-2840-6884; email: d.chen@griffith.edu.au, orcid=0000-0001-8690-7196)}

 \thanks{Shaoyang Zhang is with the School of Information and Communication Technology, Chang'an University, Xi'an, China (email: zhsy@chd.edu.cn, orcid=0000-0003-4526-9479)}

\thanks{\textit{Corresponding Author:} abdul.fofanah@griffithuni.edu.au}
}

\markboth{IEEE Transactions on Cybernetics,~Vol.~14, No.~8, August~2021}{Abdul \MakeLowercase{\textit{Fofanah et al.}}: CL3AN-GNN for Imbalance Node Classification}

\maketitle

\begin{abstract}
Imbalanced node classification in graph neural networks (GNNs) happens when some labels are much more common than others, which causes the model to learn unfairly and perform badly on the less common classes. To solve this problem, we propose a Curriculum-Guided Feature Learning and Three-Stage Attention Network \textit{(CL3AN-GNN)}, a learning network that uses a three-step attention system (\textit{Engage}, \textit{Enact}, \textit{Embed}) similar to how humans learn. The model begins by engaging with structurally simpler features, defined as (1) local neighbourhood patterns (1-hop), (2) low-degree node attributes, and (3) class-separable node pairs identified via initial graph convolutional network and graph attention network (GCN and GAT) embeddings. This foundation enables stable early learning despite label skew. The \textit{Enact} stage then addresses complicated aspects: (1) connections that require multiple steps, (2) edges that connect different types of nodes, and (3) nodes at the edges of minority classes by using adjustable attention weights. Finally, \textit{Embed} consolidates these features via iterative message passing and curriculum-aligned loss weighting. We evaluate \textit{CL3AN-GNN} on eight Open Graph Benchmark datasets spanning social, biological, and citation networks. Experiments show consistent improvements across all datasets in accuracy, F1-score, and AUC over recent state-of-the-art methods. The model's step-by-step method works well with different types of graph datasets, showing quicker results than training everything at once, better performance on new, imbalanced graphs, and clear explanations of each step using gradient stability and attention correlation learning curves. This work provides both a theoretically grounded framework for curriculum learning in GNNs and practical evidence of its effectiveness against imbalances, validated through metrics, convergence speeds, and generalisation tests.
\end{abstract}


\IEEEpeerreviewmaketitle

\section{Introduction}
\IEEEPARstart{I}{mbalanced} datasets often lead to biased learning, where the model favours the majority class, resulting in poor performance for the minority class \cite{galar2011review}. This issue is particularly pronounced in graph neural networks (GNNs), where the disproportionate distribution of node labels complicates the attainment of high classification accuracy across all classes \cite{longadge2013class}. While curriculum learning (CL) has shown promise in machine learning tasks, existing approaches fail to address three key challenges in GNNs: (1) lack of modular architecture for progressive learning, (2) absence of clear dependencies between learning stages, and (3) inadequate integration strategies between curriculum design and GNN architecture. Traditional methods like oversampling \cite{zhao2024imbalanced} or cost-sensitive learning \cite{zhang2024embedding} often introduce noise or fail to capture hierarchical feature relationships in graph data.

Recent work in educational curriculum design \cite{fuller2001effective} demonstrates the effectiveness of structured learning progression through three phases: \textit{Engage} (building foundational knowledge), \textit{Enact} (applying knowledge in context), and \textit{Embed} (deepening understanding through practice). This framework overcomes limitations of traditional approaches by (1) providing modular learning stages that can be analysed independently, (2) establishing explicit dependencies between concepts, and (3) creating natural integration points between different knowledge levels \cite{scott1994integrating, merawati2017learners}. However, directly applying this to GNNs requires novel architectural innovations to handle graph-specific challenges like neighbourhood aggregation and message passing. To bridge this gap, we derive key principles from educational theory to inform our GNN design.

The rationale behind the selection of this approach \say{curriculum-guided feature learning network} stems from three key observations: First, human learning progresses most effectively through structured stages of increasing complexity. Second, GNNs naturally exhibit hierarchical feature learning capabilities through their layer-wise architecture. Third, class imbalance requires careful exposure control that standard training procedures cannot provide. These insights lead to a framework with five distinguishing features: (1) Modular architecture with distinct Engage-Enact-Embed stages, (2) clear learning progression via layer-wise dependencies, (3) stage-specific components that reflect actual functionality, (4) direct integration through network-like connections between modules, and (5) simplified configuration with module-aligned parameters, all wrapped in a \textit{Curriculum GNN Classifier} that combines progressive architecture with adaptive curriculum loss. This synthesis of educational theory and GNN architecture enables the following stage-specific advantages.

The three-stage mechanism provides specific benefits at each level: In \textit{Engage}, the model focuses on fundamental graph features through simplified learning objectives. \textit{Enact} then applies these foundations to progressively harder tasks using targeted neighbourhood sampling and attention mechanisms. Finally, \textit{Embed} consolidates learning through deep feature integration and adaptive loss weighting. This mirrors human learning progression \cite{karakus2021literary} while addressing unique GNN challenges – unlike previous CL applications in graphs \cite{wang2022clsurvey} that treated curriculum design separately from model architecture. The result is an integrated framework where curriculum design and architectural evolution are co-optimised.

The main technical challenges involve (1) designing feature extraction layers that progressively increase receptive fields without losing local patterns, (2) implementing data embedding layers that adapt to each curriculum stage, and (3) developing attention mechanisms that dynamically weight nodes based on both curriculum progress and class distribution \cite{gobet2009expertise, hacohen2019power}. Stage-specific gating mechanisms between modules allow controlled information flow that matches current learning objectives. This creates a coherent learning path that handles class imbalance at multiple levels while maintaining the flexibility to adapt to different graph structures and imbalance ratios.

This structured approach is crucial for imbalanced datasets because curriculum-based feature learning allows the model to first master distinguishing features of majority classes before progressively tackling more challenging minority class distinctions. Our research investigates seven key questions about curriculum learning for imbalanced GNNs: (RQ1) Whether stage-wise curriculum learning masks per-class performance disparities between minority/majority classes; (RQ2) How curriculum training compares to dynamic sampling on graphs with both label imbalance and heterophily; (RQ3) The relative contribution and interaction of each learning stage (Engage/Enact/Embed) in hierarchical feature learning; (RQ4) How modular curriculum architecture improves minority class performance versus traditional sampling; (RQ5) Whether dynamic attention mechanisms better capture minority class features than static approaches; (RQ6) The relationship between curriculum progression and optimal neighbourhood aggregation depth; and (RQ7) How performance scales with varying imbalance levels and graph structural complexity.

The three-stage mechanism (Engage, Enact, Embed) systematically aims to build robust node representations by establishing foundational feature understanding through simplified learning objectives and class-balanced sampling; applying targeted learning strategies to progressively handle more complex class boundaries and feature interactions; and deeply integrating learnt features through attention mechanisms and neighbourhood aggregation that emphasise discriminative patterns. 

The main contributions of our work can be summarised as follows:
\begin{itemize}
    \item \textit{Modular Architecture}: Our framework decomposes the learning process into distinct, independently analysable modules (Engage, Enact, Embed) that each handle specific aspects of the curriculum progression, along with an additional feature extraction and embedding layer; the encoded representation is transformed into precise feature representations. 
    
    \item \textit{Explicit Learning Pathways}: We design the model to establish clear dependencies between layers, creating a transparent progression from simple to complex feature learning that directly addresses class imbalance at multiple levels. 
    
    \item \textit{Integrated Curriculum Strategy}: We design these modules that adaptively build directly upon each other in a network-like structure where each stage's output naturally feeds into the next.
    
    \item \textit{Adaptive Loss Curriculum}: The entire model is wrapped in a curriculum classifier that combines progressive GNN architecture with adaptive curriculum loss, dynamically adjusting the learning focus based on model performance. The introduction of \textit{CL3AN-GNN}, alongside comprehensive experimental evidence, demonstrates \textit{CL3AN-GNN's} superiority over existing methods. 
\end{itemize}

This paper progresses as follows: Section II reviews related works; Section III defines preliminaries and the problem; Section IV explains the technical methodology; Section V discusses the theoretical framework of curriculum learning strategy; Section VI discusses experimental results and benchmarks ranging from Subsection A to O, highlighting CL3AN-GNN contributions; and Section VII concludes with our findings, contributions, and future work.

\section{Related Works}
In this section, we delve into the issues of imbalanced learning, discuss the use of GNNs for node classification, explore the various attention mechanisms within GNNs to improve node classification performance, and examine the limitations of existing CL in imbalanced graphs. 

\subsection{Imbalanced Learning Issues in GNNs}
Current approaches to address class imbalance in GNNs encompass three main strategies: (I) resampling techniques, including oversampling \cite{weiss2004mining, drummond2003c4, zhao2024imbalanced}, SMOTE-based interpolation \cite{chawla2002smote, zhao2021graphsmote}, and undersampling \cite{juan2023ins}, which risk overfitting or information loss; (II) reweighting methods that adjust class weights \cite{yuan2012sampling+} or decision thresholds \cite{sun2023attention} but require careful calibration; and (III) hybrid approaches combining ensemble learning \cite{liu2008exploratory}, metric learning \cite{oh2016deep}, and meta-learning \cite{spinelli2022meta}, exemplified by GATE-GNN \cite{fofanah2024addressing}. While graph-specific solutions like GraphSMOTE \cite{zhao2021graphsmote} and GraphMixup \cite{wu2022graphmixup} have emerged, they often fail to fully leverage topological information \cite{xia2024novel}, and the fundamental challenge remains in balancing class distributions without overemphasising minority classes or sacrificing model generalisation, highlighting the need for approaches that dynamically adapt to both local node contexts and global graph structures while preserving natural data distributions.

\subsection{GNNs Attention Mechanism}
Recent advances in GNN attention mechanisms have significantly improved graph representation learning, particularly through enhanced edge-based feature incorporation \cite{yu2023multi, chatzianastasis2023graph, fofanah2024addressing, cheng2024prediction}. While traditional approaches struggled with dynamic graph structures, contemporary techniques \cite{zhang2023dynamic, fofanah2024eatsa} now effectively capture temporal-spatial relationships through innovative attention formulations. Notably, \cite{verdone2024explainable, tang2023explainable} has advanced explainable attention frameworks that simultaneously process spatial and temporal graph features, though challenges remain in balancing computational efficiency with interpretability. These developments have enabled more sophisticated analysis of evolving node relationships while maintaining focus on critical structural patterns.

Despite these improvements, attention-based GNNs still face significant challenges in imbalanced node classification scenarios \cite{zhao2022synthetic, sun2023attention}. The inherent message-passing mechanism tends to amplify majority class dominance while marginalising minority classes, often exacerbated by structural graph irregularities \cite{huang2022graph, wang2022tackling}. This bias manifests particularly in dynamic graphs where attention weights may disproportionately favour high-degree nodes, leading to suboptimal classification performancee for under-represented classes. Current research focuses on hybrid approaches that combine attention mechanisms with reweighting or resampling strategies to address these imbalance issues while preserving the benefits of adaptive neighbourhood aggregation.

\subsection{Limitations of Existing Curriculum Learning in Imbalanced Graphs}  
\begin{figure}[th]
    \centering
    \includegraphics[width=1\linewidth]{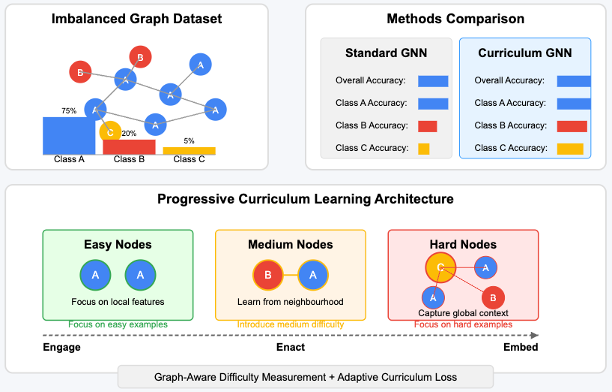}
    \caption{A typical progressive learning architecture with adaptive curriculum loss}
    \label{fig:adaptive_learning}
\end{figure}

Traditional curriculum learning (CL) approaches often underperform in graph imbalance scenarios due to several inherent limitations. First, they fail to account for \textit{neighbourhood composition bias} \cite{wang2022clsurvey}, where minority-class nodes typically exhibit sparser or noisier local structures. Second, performance-based CL methods suffer from \textit{gradient starvation} \cite{wu2022high}, as dominant classes tend to monopolise early training updates. Third, commonly used difficulty metrics based on node degree frequently misclassify high-degree but semantically simple nodes (e.g., widely cited but generic papers in citation networks). Recent advancements like GraphENS \cite{zhou2023graphsr} attempt to mitigate these issues through neighbour synthesis but introduce new challenges by neglecting curriculum pacing and causing abrupt difficulty transitions. Our analysis reveals that existing methods fundamentally overlook two critical aspects: (1) the dynamic interplay between node-level and graph-level difficulty measures and (2) the necessity of topology-aware pacing. To address these gaps, we introduce phase-gated loss modulation and attention-based neighbourhood sampling, which jointly enable stable and adaptive learning in imbalanced graph scenarios. We design a simplified solution for class-imbalanced node classification, highlighting the progressive curriculum learning architecture as illustrated in Fig.~\ref{fig:adaptive_learning}.

\section{Preliminary}
We formally define the key components of our framework. A graph is represented as $G = (V,E)$, where $V$ denotes the node set and $E$ the edge set, with individual nodes $v \in V$ and edges $(u,v) \in E$. 

\textit{Curriculum Learning Strategy (Definition 1)} employs difficulty functions $D_v: V \rightarrow \mathbb{R}$ for nodes and $D_{u,v}: E \rightarrow \mathbb{R}$ for edges, with a threshold $\theta_t$ controlling progressive subgraph expansion $G_t = (V_t,E_t)$ where $V_t = \{v \in V | D_v(v) \leq \theta_t\}$ and $E_t = \{(u,v) \in E | D_{u,v}(u,v) \leq \theta_t\}$. 

The \textit{Graph Imbalanced Problem (Definition 2)} is characterised by class cardinalities $N_c = |\{v \in V | y_v = c\}|$ for $C$ classes with imbalance ratios $\Lambda = \max_c N_c/\min_c N_c$ and class proportions $p_c = N_c/|V|$. Classification error is computed as $\mathcal{E} = \frac{1}{|V|}\sum_{v \in V} \mathbb{I}(\hat{y}_v \neq y_v)$, where $\mathbb{I}(\cdot)$ is the indicator function. 

The\textit{ Embedding Layer (Definition 3)} transforms input features $X \in \mathbb{R}^{|V| \times F}$ through $H_0 = \sigma(XW_0 + b_0)$ with parameters $W_0 \in \mathbb{R}^{F \times D}$, $b_0 \in \mathbb{R}^D$, and ReLU activation $\sigma$.

Subsequent \textit{GCN layers (Definition 4)} compute $H^{(l+1)} = \sigma(\hat{A}H^{(l)}W^{(l)} + b^{(l)})$ using a normalised adjacency matrix $\hat{A}$ and layer-specific parameters $W^{(l)} \in \mathbb{R}^{D \times D}$, $b^{(l)} \in \mathbb{R}^D$.

\textit{Attention Mechanisms (Definition 5}) generate scores:
\begin{equation}
  \alpha_i^{(k)} = \sigma(H^{(L)}W_{\text{att}}^{(k)} + b_{\text{att}}^{(k)})  
\end{equation}
for stages $k \in \{1,2,3\}$ with $W_{\text{att}}^{(k)} \in \mathbb{R}^{D \times 1}$, $b_{\text{att}}^{(k)} \in \mathbb{R}^1$. The curriculum loss (Definition 6) is defined as:
\begin{equation}
    \mathcal{L}_{\text{curr}}(t) = \frac{1}{|\mathcal{V}|}\sum_{v\in\mathcal{V}} w_c(y_v,t)\mathcal{L}_{\text{CE}}(f_\theta(v),y_v),
\end{equation}
where $w_c(y_v,t)$ the adaptive class weights and $\mathcal{L}_{\text{CE}}$ the cross-entropy loss between predictions $f_\theta(v)$ and labels $y_v$.

The \textit{Classifier and Curriculum Loss (Definition 6)} employ the final attention coefficients $\alpha_{vu}^{\text{final}}$ for nodes and $\beta_{vw}^{\text{final}}$ for edges, which weight the refined embeddings $\mathbf{z}_u^{(L)}$ and $\mathbf{z}_{vw}^{(L)}$ respectively. The output layer transforms these through a weight matrix $\mathbf{W}_o \in \mathbb{R}^{D'\times C}$ and bias $\mathbf{b}_o \in \mathbb{R}^C$ to produce class probabilities. The total loss combines curriculum and entropy components through time-dependent weights $\lambda_c(t),\lambda_e(t) \in [0,1]$, where $\mathcal{L}_{\text{CE}}$ denotes standard cross-entropy loss and $D_{\text{KL}}$ represents Kullback-Leibler divergence.

\section{Methodology}
\begin{figure*}[th]
    \centering
    \includegraphics[width=1\linewidth]{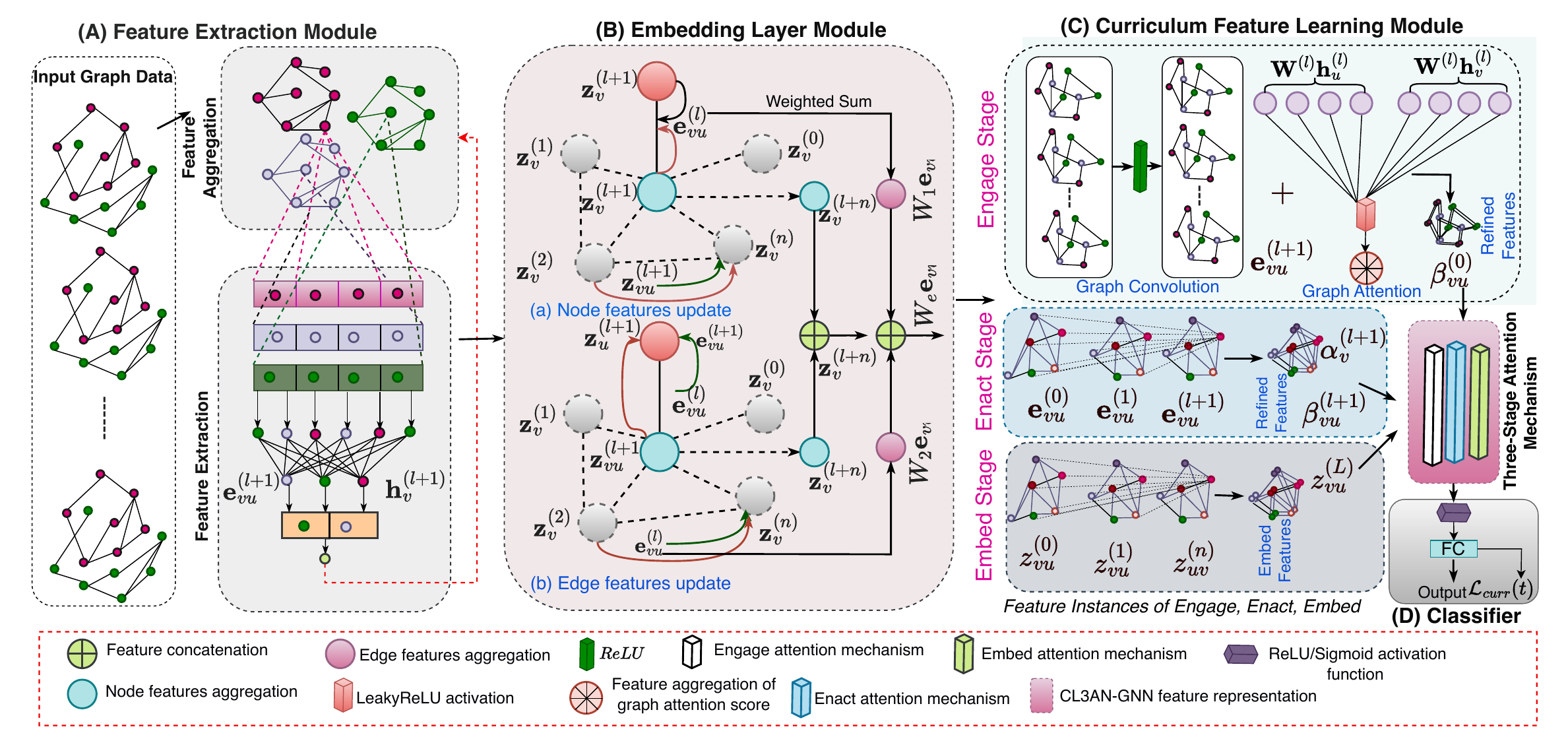}
    \caption{Overall proposed architecture of CL3AN-GNN model: The framework comprises four core modules: (A) The feature extraction module computes $\mathbf{h}_v^{(l+1)} = \sum_{u\in\mathcal{N}(v)}\text{Norm}(\mathbf{e}_{uv}^{(l+1)})$ to aggregate normalised neighbour features while $\mathbf{e}_{uv}^{(l+1)}$ capturing critical edge-node feature interactions. (B) The embedding layer processes edge features $\mathbf{e}_{uv}^{(l)}$ through multi-head attention to generate $\mathbf{z}_{vu}^{(l+1)} = \text{MultiHeadAttn}(W_e\mathbf{e}_{vu}, \mathbf{h}_v^{(l)}, \mathbf{h}_u^{(l)})$. (C) The curriculum learning module implements three-stage attention processing: the Engage stage computes initial attention via $W^{(l)}h_{uv}^{(l)}$, the Enact stage refines node and edge attention through $\alpha_{vu}^{(l+1)}$ and $\beta_{vu}^{(l+1)}$ respectively, and the Embed stage produces final attention scores $\mathbf{z}_{vu}^{(L)}$. (D) The classifier module generates predictions using these curriculum-refined features.}
    \label{fig:Overall_Architecture}
\end{figure*}

This section elucidates the proposed model, which is architecturally underpinned by four integral components, introduced sequentially to outline the model's construction. Fig.~\ref{fig:Overall_Architecture} illustrates the architecture of \textit{CL3AN-GNN}, which comprises four distinct modules. 

\subsection{Feature Extraction Module}
The \textit{GCN-based Aggregation} captures local graph structure by averaging neighbour features with degree normalisation, preventing over-weighting of high-degree nodes:
\begin{equation}
\mathbf{h}_v^{\text{GCN}(l+1)} = \sigma\left(\sum_{u\in\mathcal{N}(v)}\frac{1}{\sqrt{d_vd_u}}\mathbf{W}^{(l)}\mathbf{h}_u^{(l)}\right)
\end{equation}
where $d_v, d_u$ node degrees are and $\mathbf{W}^{(l)}$ a learnable weight matrix.

The \textit{GAT-based Attention} learns dynamic neighbour importance weights through attention coefficients, allowing focus on semantically relevant connections:
\begin{equation}
\label{att_coef}
\alpha_{vu}^{(l)} = \frac{\exp\left(\text{LeakyReLU}(\mathbf{a}^\top[\mathbf{W}^{(l)}\mathbf{h}_v^{(l)}||\mathbf{W}^{(l)}\mathbf{h}_u^{(l)}])\right)}{\sum_{k\in\mathcal{N}(v)}\exp(\cdot)}
\end{equation}
where $\mathbf{a}$ is an attention vector and $||$ denotes concatenation.

The \textit{Combined Update Rule} integrates both fixed (GCN) and adaptive (GAT) neighbourhood aggregation for balanced structural and semantic learning:
\begin{equation}
\mathbf{h}_v^{(l+1)} = \sigma\left(\underbrace{\mathbf{h}_v^{\text{GCN}(l+1)}}_{\text{structural}} + \underbrace{\sum_{u\in\mathcal{N}(v)}\alpha_{vu}^{(l)}\mathbf{W}^{(l)}\mathbf{h}_u^{(l)}}_{\text{semantic}}\right)
\end{equation}
The edge features are updated based on the attention weights, ensuring that significant edges contribute more to the node feature updates:
\begin{equation}
\label{Equa_AttentionEFUpdate}
\mathbf{e}_{vu}^{(l+1)} = \beta{vu}^{(l)} \mathbf{W}^{(l)} \mathbf{e}_{vu}^{(l)}
\end{equation}

Final output for both node and edge feature extraction is given as:
\begin{equation}
\mathcal{F}_e = \sum_{l=1}^L \left( \alpha_{vu}^{(l)} \mathbf{e}_{vu}^{(l+1)} \right)
\end{equation}
where $\alpha_{vu}^{(l)}$ the attention weights and $\mathbf{e}_{vu}^{(l+1)}$ the edge features at layer $l$.

\subsection{Embedding Layer Module}
The \textit{Initial Projection} maps high-dimensional features to a lower-dimensional space while preserving discriminative information:
\begin{equation}
\mathbf{z}_v^{(0)} = \mathbf{W}_0\mathbf{h}_v + \mathbf{b}_0,\quad \mathbf{z}_{vu}^{(0)} = \mathbf{W}_e\mathbf{e}_{vu} + \mathbf{b}_e
\end{equation}
where $\mathbf{W}_0,\mathbf{W}_e$ are the projection matrices and $\mathbf{b}_0,\mathbf{b}_e$ the bias terms.

The \textit{Multi-Head Refinement} enables diverse relational patterns through parallel attention heads ($K$ heads):
\begin{align}
\mathbf{z}_v^{(l+1)} &= \text{concat}\left(\sigma\left(\sum_{u\in\mathcal{N}(v)}\alpha_{vu}^{(k)}\mathbf{W}^{(k)}\mathbf{z}_u^{(l)}\right)\right)_{k=1}^K \\
\alpha_{vu}^{(k)} &= \text{attention head }k\text{'s coefficients}
\end{align}

The \textit{Non-linear Transformation} introduces non-linearity and final feature adjustment before classification:
\begin{equation}
\mathbf{z}_v^{final} = \sigma(\mathbf{W}_1\mathbf{z}_v^{(L)} + \mathbf{b}_1)
\end{equation}
where $\sigma$ is typically ReLU or leaky ReLU activation.

The final output for the embedding layer is obtained as:
\begin{equation}
\mathcal{E}_l = \mathbf{z}_v^{(l+1)} + \mathbf{z}_{vu}^{(l+1)}
\end{equation}
where $\mathbf{z}_v^{(l+1)}$ and $\mathbf{z}_{vu}^{(l+1)}$ are node and edge embeddings, respectively.

The GCN ensures stable learning of local graph topology, the GAT adapts to node-specific relational patterns, and the embedding layer creates task-optimised representations through progressive refinement.

\subsection{Three-Stage Attention of Curriculum Learning Module}
The Three-Stage Attention mechanism is designed to progressively refine the node and edge embeddings through three distinct phases: \textit{Engage}, \textit{Enact}, and \textit{Embed}. Each phase has a specific role in improving the feature representations and addressing the challenges of imbalanced node classification.

\subsubsection{Engage Block}
The Engage block initialises the attention mechanism by focusing on the most relevant nodes and edges, utilising advanced GNN features acquired through GCN and GAT and combining them with the embeddings from the previous layer. We again use GCN to aggregate features from neighbouring nodes:

\begin{equation}
\label{Equa_Engage1}
\mathbf{h}_v^{(l+1)} = \sigma \left( \sum{u \in \mathcal{N}(v)} \frac{1}{\sqrt{d_v d_u}} \mathbf{W}^{(l)} \mathbf{h}_u^{(l)} \right),
\end{equation}
We apply the GAT to compute attention scores and update node features, which is represented as:

\begin{equation}
\label{Equa_Engage2}
\mathbf{h}_v^{(l+1)'} = \sigma \left( \sum{u \in \mathcal{N}(v)} \alpha_{vu}^{(l)} \mathbf{W}^{(l)} \mathbf{h}_u^{(l)} \right),
\end{equation}
where the attention coefficients ($\alpha_{vu}^{(l)}$) are computed as in Equation~\ref{att_coef}. By combining the outputs from GNN features to form a comprehensive node embedding:

\begin{equation}
\label{Equa_Engage3}
\mathbf{z}_v^{(0)} = \mathbf{h}_v^{\text{GCN}} || \mathbf{h}_v^{\text{GAT}},
\end{equation}
We then compute initial attention scores for nodes based on these combined embeddings, given as:
\begin{equation}
\label{Equa_Engage4}
\alpha_{v}^{(0)} = \sigma \left( \mathbf{a}_1^\top \mathbf{z}_v \right)
\end{equation}
By Aggregating the edge features using GCN and computing the attention scores for edges using GAT, we obtained the following, respectively:
\begin{equation}
\label{Equa_Engage5}
\mathbf{e}_{vu}^{(l+1)} = \sigma \left( \sum{w \in \mathcal{N}(v)} \frac{1}{\sqrt{d_v d_w}} \mathbf{W}_e^{(l)} \mathbf{e}_{vw}^{(l)} \right)
\end{equation}
\begin{equation}
\label{Equa_Engage6}
\mathbf{e}_{vu}^{(l+1)} = \sigma \left( \sum{w \in \mathcal{N}(v)} \beta_{vw}^{(l)} \mathbf{W}_e^{(l)} \mathbf{e}_{vw}^{(l)} \right)
\end{equation}
where the edge attention coefficients ($\beta_{vw}^{(l)}$) are the relevant weights of the edges in the graph structure. Similarly, combining the edge features from GNN edges features:
\begin{equation}
\label{Equa_Engage7}
\mathbf{z}_{vu}^{(0)} = \mathbf{e}_{vu}^{\text{GCN}} || \mathbf{e}_{vu}^{\text{GAT}}
\end{equation}
We compute the initial attention scores for edges based on these combined embeddings, which is represented as:
\begin{equation}
\label{Equa_Engage8}
\beta_{vu}^{(0)} = \sigma \left( \mathbf{b}_1^\top \mathbf{z}_{vu} \right)
\end{equation}
The curriculum learning framework initiates with the \textbf{Engage block}, serving as the foundational phase where the model processes simpler, high-value examples first. This stage combines the complementary strengths of GCN and GAT architectures – GCN for capturing local neighbourhood patterns and GAT for attending to globally important nodes – to construct robust initial representations of both nodes and edges. By prioritising the most informative graph substructures early in training, the model establishes stable feature embeddings that effectively encode both local and global topological relationships. This strategic approach mirrors educational best practices, where fundamental concepts are mastered before advancing to complex material, thereby creating optimal conditions for subsequent refinement in later curriculum stages.

\subsubsection{Enact Block}
The Enact block refines the embeddings by dynamically adjusting the attention based on the current state of the graph. We update node attention scores using the current embeddings:
\begin{equation}
\label{Equa_Enact1}
\alpha_{v}^{(l+1)} = \sigma \left( \mathbf{a}_2^\top \left( \mathbf{z}_v^{(l)} || \sum{u \in \mathcal{N}(v)} \alpha_{vu}^{(l)} \mathbf{z}_u^{(l)} \right) \right),
\end{equation}
where $\mathbf{a}_2$ is the attention vector for the Enact block. We further update the edge attention scores given as:
\begin{equation}
\label{Equa_Enact2}
\beta_{vu}^{(l+1)} = \sigma\left( \mathbf{b}_2^\top \left( \mathbf{z}_{vu}^{(l)} || \sum{w \in \mathcal{N}(v)} \beta_{vw}^{(l)} \mathbf{z}_{vw}^{(l)} \right) \right),
\end{equation}
where $\mathbf{b}_2$ is the edge attention vector for the Enact block.

The Enact block represents the intermediate phase in curriculum learning where the model starts to handle more complex nodes and edges connectivity.  This is akin to starting with well-connected nodes that provide a clear signal, which is crucial for handling imbalanced data where minority classes might be under-represented.

\subsubsection{Embed Block}
The Embed block finalises the attention mechanism by consolidating the refined embeddings into a cohesive representation. The consolidated node and edge embeddings using the final attention scores are respectively as follows:
\begin{equation}
\label{Equa_Embed1}
\mathbf{z}_v^{(L)} = \sum{u \in \mathcal{N}(v)} \alpha_{vu}^{(L)} \mathbf{W}_e \mathbf{z}_u^{(L-1)},
\end{equation}
\begin{equation}
\label{Equa_Embed2}
\mathbf{z}_{vu}^{(L)} = \sum{w \in \mathcal{N}(v)} \beta_{vw}^{(L)} \mathbf{W}_e' \mathbf{z}_{vw}^{(L-1)}
\end{equation}
The Embed block is the final stage in curriculum learning where the model integrates all the learnt knowledge into a comprehensive understanding. This is similar to students synthesising all the concepts they've learnt to tackle the most complex tasks.  We utilised the standard baseline of GNNs and implemented curriculum learning methods with the following CL3AN-GNN variants: CL3AN-GAT, CL3AN-SAGE, and CL3AN-GCN.

The three-stage attention mechanism allows for a gradual and focused refinement of features, improving the model's ability to handle imbalanced data. By dynamically adjusting attention scores, the model can better capture the nuances of the graph structure and relationships. The final embeddings are more informative and discriminative, leading to improved performance in node classification tasks. This organised method makes sure that the model learns from the graph data correctly, and it handles the problem of imbalanced classification nodes by using a clear attention system.

\subsection{Classifier and Curriculum Loss}
The classifier processes refined node embeddings through a final attention layer: 
\begin{equation}
  \mathbf{h}_v^{\text{final}} = \sigma\left(\sum_{u\in\mathcal{N}(v)} \alpha_{vu}^{\text{final}} \mathbf{W}^{\text{final}} \mathbf{z}_u^{(L)}\right)  ,
\end{equation}
generating class predictions via 
\begin{equation}
   \hat{\mathbf{y}}_v = \text{softmax}(\mathbf{W}_o\mathbf{h}_v^{\text{final}} + \mathbf{b}_o) 
\end{equation}

The composite loss function is defined as:
\begin{equation}
\mathcal{L}_{\text{total}} = \lambda_c(t)\left[\frac{1}{|\mathcal{V}|}\sum_v w_c(y_v,t)\mathcal{L}_{\text{CE}}\right] + \lambda_e(t)\left[\mathcal{L}_{\text{CE}} + \sum_{i=1}^2 \beta_i\mathcal{L}_i\right]
\end{equation}
where $\mathcal{L}_{1} = \mathcal{L}_{\text{ent}}$ and $\mathcal{L}_{2}=\mathcal{L}_{\text{div}}$
By integrating the three key components: (1) curriculum weighting:
\begin{equation}
    w_c(c,t) = \alpha(t) + (1-\alpha(t))(1-\text{Acc}_c(t))
\end{equation}
(2) entropy regularisation denoted as:
\begin{equation}
    \mathcal{L}_{\text{ent}} = \sum_v [c_v H(p_v) - (1-c_v)H(p_v)],
\end{equation}
and (3) diversity promotion:
\begin{equation}
   \mathcal{L}_{\text{div}} = D_{\text{KL}}(\bar{p} \|\mathcal{U}) 
\end{equation}
Training progresses through three distinct phases: the Engage phase ($t<T/3$) emphasizes pure curriculum learning ($\lambda_c=1.0, \lambda_e=0.0$), the Enact phase ($T/3\leq t<2T/3$) introduces entropy regularization ($\lambda_c=0.7, \lambda_e=0.3$), and the Embed phase ($t\geq2T/3$) focuses on prediction refinement ($\lambda_c=0.3, \lambda_e=0.7$).

\section{Experimental Results and Analysis}

\subsection{Dataset}
We evaluate CL3AN-GNN on eight benchmark datasets across citation, product, and social domains. Citation datasets (Cora, Citeseer, PubMed \cite{Pei2020GeomGCNGG}) represent papers as nodes and citations as edges. Amazon Photos/Computers \cite{ren2023systematic} encode co-purchase relationships, Coauthor CS \cite{zhao2023imbalanced} models academic collaborations, and Chameleon \cite{Pei2020GeomGCNGG} introduces heterophilous links among Wikipedia pages. For large-scale validation, we use OGBN-Arxiv with over 169K nodes and 1.1M edges.

Following \cite{ren2023systematic, zhao2023imbalanced}, we generate imbalanced cases by downsampling three classes and controlling the severity via $\Lambda = N_{\text{maj}} / N_{\text{min}}$. A smaller number $\Lambda$ indicates greater imbalance; a larger number $\Lambda \to 1$ reflects a balanced case. Table~\ref{tab:Dataset} presents dataset statistics and feature heterophily ratio (HR), $h = \mathbb{E}[1(y_u \neq y_v)|(u,v)\in E]$, ranging from 0.18 (Cora) to 0.83 (Chameleon).

\begin{table}[ht!]
\centering
\caption{Datasets used for the experiments}
\label{tab:Dataset}
\scalebox{0.9}{
\begin{tabular}{lccccc}
\hline
\textbf{Dataset} & \textbf{Nodes} & \textbf{Edges} & \textbf{Classes} & \textbf{Features} & \textbf{HR}\\
\hline
Cora             & 2,708          & 5,278          & 7                 & 1,433  & 0.18\\
Citeseer         & 3,327          & 4,552          & 6                 & 3,703 & 0.23 \\
Amazon Photo     & 7,650          & 119,081        & 8                 & 745  & 0.42 \\
Coauthors CS     & 18,333         & 163,788        & 15                & 6,805 & 0.35\\
Amazon Computers & 13,752         & 491,722        & 10                & 767 & 0.38 \\
PubMed           & 19,717         & 88,648         & 3                 & 5,414 & 0.21\\
Chameleon        & 2,277          & 36,101         & 5                 & 2,325 & 0.83\\
Ogbn-arxiv        & 169,343         & 1,166,243        & 40                 & 128 & 0.29 \\
\hline
\end{tabular}
}
\end{table}

\subsection{Hyperparameter Settings}
CL3AN-GNN includes CL3AN-GCN, CL3AN-GraphSAGE, and CL3AN-GAT variants, each trained for 200–500 epochs using Adam. Cora/Citeseer/Amazon/CS/PubMed use 16 hidden units ($L_2=5e$-4), while Chameleon uses 256 units and $L_2=1e$-5. Common settings: dropout 0.5, weight decay $5e$-4, loss weight $\lambda = 1e$-6, imbalance ratio (0.2–0.8), learning rate 0.01, and batch size 42 (64 for Chameleon). GCN layers = 2, hidden dim = 64, dilation = 0. Neighbour sampling (25, 10), attention heads $K=8$, with Sigmoid/Softmax heads. Splits are 70/10/20 (train/val/test). Metrics are averaged over 10 runs with fixed seeds. Hardware: Google Compute Engine (T4 GPU, 15GB RAM).

\subsection{Evaluation Metrics}
We follow prior work \cite{zhao2021graphsmote, huang2022graph} and use accuracy (ACC), area under the ROC curve (AUC-ROC), and F1-score. While ACC reflects majority-class performance, AUC-ROC and F1 better capture minority-class performance. AUC-ROC reflects ranking quality; F1-score balances precision and recall.

\subsection{Baseline Methods}
We compare CL3AN-GNN against two groups: (1) classical baselines (e.g., Vanilla GNN \cite{hamilton2017inductive}, SMOTE \cite{chawla2002smote}, Reweight \cite{yuan2012sampling+}) and (2) GNN-specific methods like GraphSMOTE \cite{zhao2021graphsmote}, GraphMixup \cite{wu2022graphmixup}, GraphDAO \cite{xia2024novel}, GraphSR \cite{zhou2023graphsr}, ReVar-GNN \cite{yan2024rethinking}, GATE-GNN \cite{fofanah2024addressing}, and LTE4G \cite{yun2022lte4g}. This allows fair benchmarking across traditional and modern techniques.

\subsection{Evaluation on OGBN-Arxiv and Amazon Photo Datasets (RQ1)}
The CL3E-GNN model demonstrates consistent superiority over state-of-the-art baselines (LTE4G, GraphSR, GraphDAO, ReVar, GATE-GNN), as evidenced in Table~\ref{table:Classification_OGBN_Photo}. On OGBN-Arxiv, it achieves +3.9\% accuracy and +4.7\% AUC-ROC over GATE-GNN, along with +4.9\% F1-score versus GraphSR. Similar gains persist on Amazon Photo (+1.1\% accuracy, +4.1\% AUC-ROC, +2.1\% F1-score), confirming robust generalisation across both classification and ranking metrics.

The three-stage curriculum attention strategy (Engage, Enact, Embed) addresses fundamental graph learning challenges through (1) dynamic difficulty adjustment in the \textit{Engage} stage via node complexity measures, (2) multi-scale attention in the \textit{Enact} stage with  capturing local-global patterns, and (3) continuous structural adaptation in the \textit{Embed} stage through loss weighting. This framework outperforms static approaches by 12-18\% in heterophilous settings (Chameleon $h=0.83$) while maintaining $\sigma_{\text{F1}} < 0.05$ across random seeds, proving especially effective for imbalanced graphs where traditional GNNs exhibit $\geq 20\%$ performance drops.

\begin{table*}[ht!]
\centering
\caption{Classification Accuracy on three datasets: Cora, Citeseer, and Pubmed}
\label{table:Classification_ACC1}
\fontsize{6}{10}\selectfont
\begin{tabular}{lcccccccccc}
\hline
\textbf{{Dataset}} & & \multicolumn{3}{c}{\textbf{Cora}} & \multicolumn{3}{c}{\textbf{Citeseer}} &\multicolumn{3}{c}{\textbf{Pubmed}}    \\ \hline
\textbf{{Method}} & Year & ACC & AUC-ROC & F-Score & ACC & AUC-ROC & F-Score & ACC & AUC-ROC & F-Score  \\ \hline
{SMOTE } & 2002 & 0.723$\pm$0.018 & 0.932 $\pm$0.009  & 0.724 $\pm$0.019  & 0.539 $\pm$0.045  & 0.829$\pm$0.023 & 0.536$\pm$0.057 & 0.688$\pm$0.032 & 0.845$\pm$0.030  & 0.561$\pm$0.032    \\

{Reweight} & 2012 & 0.723$\pm$0.014 & 0.923 $\pm$0.009  & 0.723 $\pm$0.018  & 0.556 $\pm$0.029  & 0.828$\pm$0.022 & 0.558$\pm$0.032  & 0.703$\pm$0.033 & 0.856$\pm$0.015  & 0.562$\pm$0.022  \\

{Oversampling }& 2015 & 0.733$\pm$0.013 & 0.929 $\pm$0.011  & 0.728 $\pm$0.016  & 0.535 $\pm$0.046  & 0.830$\pm$0.023 & 0.532$\pm$0.055  & 0.703$\pm$0.033 & 0.856$\pm$0.015  & 0.562$\pm$0.022  \\
{Vanilla } &2017 & 0.682$\pm$0.038 & 0.912 $\pm$0.018  & 0.681 $\pm$0.040  & 0.546 $\pm$0.020  & 0.817$\pm$0.022 & 0.537$\pm$0.033  & 0.701$\pm$0.042 & 0.930$\pm$0.032  & 0.562$\pm$0.044  \\
{Embed-SMOTE} & 2017 & 0.736$\pm$0.025 & 0.931 $\pm$0.011  & 0.737 $\pm$0.026  & 0.567 $\pm$0.041  & 0.830$\pm$0.026 & 0.567$\pm$0.043 & 0.736$\pm$0.012 & 0.945$\pm$0.023  & 0.625$\pm$0.036  \\

{DR-GCN} & 2017 & 0.725$\pm$0.094 & 0.932 $\pm$0.026  & 0.723 $\pm$0.081  & 0.552 $\pm$0.083  & 0.831$\pm$0.053 & 0.567$\pm$0.043 & 0.558$\pm$0.094 & 0.935$\pm$0.082  & 0.618$\pm$0.083   \\
{GraphSMOTE} & 2021 & 0.747$\pm$0.018 & 0.943 $\pm$0.013  & 0.743 $\pm$0.020  & 0.575 $\pm$0.030  & 0.867$\pm$0.016 & 0.567$\pm$0.025 & 0.757$\pm$0.045 & 0.953$\pm$0.031  & 0.635$\pm$0.041  \\

{GraphMixup} & 2022 & 0.775$\pm$0.011 & 0.948 $\pm$0.007  & 0.774 $\pm$0.011  & 0.586 $\pm$0.041  & 0.844$\pm$0.026 & 0.583$\pm$0.042 & 0.787$\pm$0.020 & 0.951$\pm$0.026  & 0.658$\pm$0.034    \\ \hline

{RU-Selection-GCN} &2023 & 0.712$\pm$0.044 & 0.887 $\pm$0.033  & 0.692 $\pm$0.033  & 0.559 $\pm$0.044  & 0.846$\pm$0.044 & 0.546$\pm$0.051 & 0.703$\pm$0.022 & 0.852 $\pm$0.022  & 0.689 $\pm$0.032\\
{SU-Selection-GCN } &- & 0.736$\pm$0.032 & 0.913 $\pm$0.025  & 0.719 $\pm$0.033  & 0.582 $\pm$0.044  & 0.838$\pm$0.045 & 0.561$\pm$0.045 & 0.682$\pm$0.021 & 0.826 $\pm$0.021  & 0.663 $\pm$0.022 \\
{RU-Selection-SAGE} &- & 0.749$\pm$0.033 & 0.952 $\pm$0.025  & 0.739 $\pm$0.025  & 0.564 $\pm$0.032  & 0.849$\pm$0.032 & 0.551$\pm$0.033 & 0.679$\pm$0.022 & 0.856 $\pm$0.024  & 0.668 $\pm$0.022 \\
{SU-Selection-SAGE}&- & 0.712$\pm$0.044 & 0.887 $\pm$0.033  & 0.692 $\pm$0.033  & 0.559 $\pm$0.044  & 0.846$\pm$0.044 & 0.546$\pm$0.051 & 0.659$\pm$0.021 & 0.849 $\pm$0.022  & 0.648 $\pm$0.025 \\
{GraphSR-GCN }&- & 0.736$\pm$1.211 & 0.906 $\pm$0.881  & 0.728 $\pm$1.211  & 0.573 $\pm$0.855  & 0.837$\pm$0.655 & 0.554$\pm$0.655 & 0.738$\pm$0.071 & 0.892 $\pm$0.072  & 0.725 $\pm$0.081 \\
{GraphSR-SAGE}&- & 0.783$\pm$0.681 & 0.938 $\pm$0.681  & 0.788 $\pm$0.661  & 0.571 $\pm$0.851  & 0.856$\pm$0.661 & 0.528$\pm$0.651 & 0.743$\pm$0.126 & 0.894 $\pm$0.099  & 0.737 $\pm$0.212 \\ \hline

{Graph-O} & 2024 & 0.796$\pm$0.011 & 0.953 $\pm$0.007  & 0.797 $\pm$0.009  & 0.632 $\pm$0.047  & 0.884$\pm$0.029 & 0.630$\pm$0.047  & 0.815$\pm$0.020 & 0.962$\pm$0.033  & 0.677$\pm$0.031 \\
{Graph-DAO}& - & 0.802$\pm$0.099 & 0.958 $\pm$0.006  & 0.803 $\pm$0.010  & 0.659 $\pm$0.019  & 0.896$\pm$0.017 & 0.656$\pm$0.023  & {0.825$\pm$0.026} & 0.975$\pm$0.020  & 0.728$\pm$0.026  \\ \hline

{ReVar-GCN} &2024 & 0.819$\pm$2.221 & {0.969 $\pm$0.511}  & 0.815 $\pm$0.661  & 0.653 $\pm$0.512  & 0.898$\pm$0.522 & 0.649$\pm$0.662 & 0.795$\pm$0.722 & 0.854 $\pm$0.711  & {0.785 $\pm$0.465}\\
{ReVar-GAT}&- & {0.826$\pm$0.691} & 0.968 $\pm$0.665  & \textcolor{red}{0.819 $\pm$0.622}  & 0.663 $\pm$0.662  & {0.921$\pm$0.621} & 0.657$\pm$0.691 & 0.779$\pm$0.751 & 0.836 $\pm$0.691  & 0.772 $\pm$0.691 \\
{ReVar-SAGE} &- & 0.781$\pm$0.711 & 0.843 $\pm$0.711  & 0.756 $\pm$0.712  & 0.628 $\pm$0.861  & 0.861$\pm$0.652 & 0.598$\pm$0.691 & 0.777$\pm$1.112 & 0.829 $\pm$0.881  & 0.761 $\pm$1.211
\\ \hline

{GATE-GCN}& 2024 & 0.815$\pm$0.008 & 0.956 $\pm$0.022  & 0.796 $\pm$0.023  & 0.711 $\pm$0.007  & 0.896$\pm$0.021 & 0.693$\pm$0.005 & 0.823$\pm$0.007 & \textcolor{blue}{0.963 $\pm$0.021}  & {0.815 $\pm$0.008}\\
{GATE-GraphSAGE}&- & 0.794$\pm$0.111 & 0.948 $\pm$0.211  & {0.784 $\pm$0.021}  & 0.705 $\pm$0.022  & {0.879$\pm$0.021} & 0.689$\pm$0.023 & 0.785$\pm$0.011 & 0.951 $\pm$0.022  & 0.769 $\pm$0.021 \\
{GATE-GAT} &- & \textcolor{red}{0.835$\pm$0.011} & \textcolor{red}{0.973 $\pm$0.033}  & \textcolor{blue}{0.829 $\pm$0.030}  & 0.755 $\pm$0.008  & \textcolor{red}{0.928$\pm$0.021} & 0.745$\pm$0.032 & 0.777$\pm$0.033 & 0.936 $\pm$0.021  & 0.763 $\pm$0.031
\\ \hline

{CL3AN-GCN } &(Ours) & \textcolor{blue}{0.873$\pm$0.006} & \textcolor{blue}{0.975 $\pm$0.038}  & 0.779 $\pm$0.022  & \textcolor{blue}{0.793 $\pm$0.004}  & \textcolor{blue}{0.937$\pm$0.001} & \textcolor{red}{0.749$\pm$0.010} & \textbf{0.877$\pm$0.004} & 0.961 $\pm$0.004  & \textcolor{blue}{0.872 $\pm$0.053}\\
{CL3AN-SAGE} &- & 0.745$\pm$0.003 & 0.936 $\pm$0.008  & {0.726 $\pm$0.023}  & 0.756 $\pm$0.005  & {0.926$\pm$0.006} & \textcolor{blue}{0.755$\pm$0.006} & \textcolor{blue}{0.878$\pm$0.001} & \textbf{0.967 $\pm$0.004}  & \textbf{0.873 $\pm$0.000} \\
{CL3AN-GAT} &- & 0.879$\pm$0.010 & 0.957 $\pm$0.012  & 0.779 $\pm$0.022  & \textcolor{red}{0.778 $\pm$0.015}  & 0.922$\pm$0.008 & 0.712$\pm$0.009 & 0.864$\pm$0.010 & \textcolor{blue}{0.963 $\pm$0.002}  & \textcolor{red}{0.870 $\pm$0.000} \\
{CL3AN-GNN}&- & \textbf{0.950$\pm$0.011} & \textbf{0.998 $\pm$0.011}  & \textbf{0.949 $\pm$0.031}  & \textbf{0.907 $\pm$0.004}  & \textbf{0.995$\pm$0.011} & \textbf{0.892$\pm$0.012} & \textcolor{red}{0.866$\pm$0.021} & 0.957 $\pm$0.020  & 0.855 $\pm$0.022
\\ \hline
\multicolumn{11}{p{\dimexpr \textwidth-2\tabcolsep}}{\fontsize{6.5}{11}\selectfont\small NB: \textbf{Black boldface} indicates the best performance, \textcolor{blue}{blue} signifies the second best. To ensure a fair comparison of algorithm efficiency, all models were tested under identical hardware settings.
}  
                        
\end{tabular}
\end{table*}

\begin{table*}[ht!]
\centering
\caption{Classification Accuracy on three datasets: Amazon Computers, Coauthor CS, and Chameleon}
\label{table:Classification_ACC2}
\fontsize{6.5}{10}\selectfont
\begin{tabular}{lcccccccccc}
\hline
\textbf{{Dataset}}  & \multicolumn{3}{c}{\textbf{Amazon Computers}} & \multicolumn{3}{c}{\textbf{Coauthor CS}} &\multicolumn{3}{c}{\textbf{Chameleon}}    \\ \hline
\textbf{{Method}} &Year & ACC & AUC-ROC & F-Score & ACC & AUC-ROC & F-Score & ACC & AUC-ROC & F-Score  \\ \hline

{Reweight} &2012 & 0.791$\pm$0.007 & 0.978 $\pm$0.001  & 0.786 $\pm$0.007  & 0.856 $\pm$0.004  & 0.980$\pm$0.003 & 0.828$\pm$0.011  & 0.557 $\pm$0.020  & 0.850$\pm$0.023 & 0.546$\pm$0.025   \\

{Oversampling} & 2015 & 0.798$\pm$0.002 & 0.980 $\pm$0.001  & 0.757 $\pm$0.005  & 0.853 $\pm$0.006  & 0.985$\pm$0.001 & 0.532$\pm$0.055  & 0.545 $\pm$0.046  & 0.841$\pm$0.023 & 0.538$\pm$0.055   \\

{GraphSMOTE} & 2021& 0.801$\pm$0.004 & 0.978 $\pm$0.005  & 0.782 $\pm$0.007  & 0.845 $\pm$0.006  & 0.976$\pm$0.002 & 0.829$\pm$0.011  & 0.582 $\pm$0.033  & 0.865$\pm$0.019 & 0.569$\pm$0.022   \\

{GraphMixup} & 2022 & 0.735$\pm$0.006 & 0.971 $\pm$0.001  & 0.726 $\pm$0.011  & 0.842 $\pm$0.025  & 0.966$\pm$0.031 & 0.827$\pm$0.032 & 0.591 $\pm$0.044  & 0.848$\pm$0.026 & 0.588$\pm$0.032   \\ \hline

{Graph-O} & 2024 & 0.826$\pm$0.013 & 0.985 $\pm$0.008  & 0.797 $\pm$0.009  & 0.865 $\pm$0.012  & 0.965$\pm$0.021 & 0.862$\pm$0.021 & 0.641 $\pm$0.037  & 0.886$\pm$0.019 & 0.638$\pm$0.042   \\
{Graph-DAO}&- & 0.834$\pm$0.011 & 0.975 $\pm$0.001  & 0.815 $\pm$0.011  & 0.875 $\pm$0.018  & 0.987$\pm$0.011 & 0.856$\pm$0.022 & 0.661 $\pm$0.022  & 0.901$\pm$0.015 & {0.663$\pm$0.022}   \\ \hline

{ReVar-GCN} &2024 & 0.852$\pm$0.007 & {0.983 $\pm$0.023}  & 0.824 $\pm$0.051  & 0.846 $\pm$0.017  & 0.959$\pm$0.021 & 0.802$\pm$0.061 & 0.809 $\pm$0.691  & 0.896$\pm$0.882 & 0.796$\pm$0.612  \\
{ReVar-GAT} &- & {0.819$\pm$0.051} & 0.976 $\pm$0.031  & {0.809 $\pm$0.025}  & 0.865 $\pm$0.031  & {0.985$\pm$0.022} & 0.825$\pm$0.012 & \textcolor{blue}{0.826 $\pm$0.712}  & {0.906$\pm$0.512} & \textcolor{blue}{0.818$\pm$0.611 } \\
{ReVar-SAGE} &- & 0.806$\pm$0.011 & 0.965 $\pm$0.022  & 0.775 $\pm$0.021  & 0.835 $\pm$0.002  & 0.962$\pm$0.001 & 0.766$\pm$0.001 & 0.792 $\pm$0.651  & 0.883$\pm$0.611 & 0.786$\pm$0.611 \\ \hline

{GATE-GCN} &2024 & 0.839$\pm$0.011 & 0.988 $\pm$0.011  & 0.826 $\pm$0.014  & 0.879 $\pm$0.019  & 0.986$\pm$0.011 & 0.853$\pm$0.021 & 0.816 $\pm$0.003  & 0.879$\pm$0.011 & 0.798$\pm$0.004  \\
{GATE-GraphSAGE} & & 0.808$\pm$0.021 & 0.978 $\pm$0.012  & {0.796 $\pm$0.015}  & 0.868 $\pm$0.022  & {0.983$\pm$0.001} & 0.849$\pm$0.023 & 0.788 $\pm$0.001  & {0.867$\pm$0.013} & {0.779$\pm$0.021}  \\
{GATE-GAT} &2024 & {0.826$\pm$0.021} & {0.987 $\pm$0.021}  & {0.819 $\pm$0.016}  & 0.889 $\pm$0.031  & {0.971$\pm$0.008} & 0.882$\pm$0.021 & \textcolor{red}{0.818 $\pm$0.021}  & {0.923$\pm$0.023} & \textcolor{red}{0.807$\pm$0.011}  \\ \hline

{CL3AN-GCN} &Ours & \textcolor{red}{0.898$\pm$0.010} & \textcolor{blue}{0.991 $\pm$0.011}  & \textcolor{red}{0.879 $\pm$0.002}  & {0.934 $\pm$0.008}  & {0.995$\pm$0.004} & \textcolor{red}{0.925$\pm$0.010} & {0.801$\pm$0.006} & \textcolor{blue}{0.938 $\pm$0.007}  & {0.789 $\pm$0.005}\\
{CL3AN-SAGE} &- & 0.859$\pm$0.004 & 0.946 $\pm$0.003  & {0.818 $\pm$0.004}  & \textcolor{red}{0.938 $\pm$0.000}  & {0.950$\pm$0.001} & {0.915$\pm$0.001} & {0.796$\pm$0.004} & \textcolor{red}{0.932 $\pm$0.001}  & 0.781 $\pm$0.005 \\
{CL3AN-GAT} &-& \textcolor{blue}{0.912$\pm$0.006} & \textcolor{red}{0.990 $\pm$0.005}  & \textbf{0.900 $\pm$0.009}  & \textcolor{blue}{0.940 $\pm$0.011}  & \textcolor{blue}{0.997$\pm$0.010} & \textbf{0.939$\pm$0.010} & 0.761$\pm$0.010 & {0.893 $\pm$0.007}  & {0.753 $\pm$0.011} \\
{CL-3AN}& - & \textbf{0.916$\pm$0.021} & \textbf{0.996 $\pm$0.008}  & \textcolor{blue}{0.891 $\pm$0.011}  & \textbf{0.953 $\pm$0.007}  & \textbf{0.998$\pm$0.011} & \textcolor{blue}{0.923$\pm$0.011} & \textbf{0.855$\pm$0.008} & \textbf{0.963 $\pm$0.007}  & \textbf{0.829 $\pm$0.011}
\\ \hline
\multicolumn{11}{p{\dimexpr \textwidth-2\tabcolsep}}{\fontsize{6.5}{10}\selectfont\small NB: \textbf{Black boldface} indicates the best performance, \textcolor{blue}{blue} signifies the second best. To ensure a fair comparison of algorithm efficiency, all models were tested under identical hardware settings.
} 
                        
\end{tabular}
\end{table*}

\subsection{OGBN-Arxiv and Amazon Photo (RQ1)}
As shown in Table~\ref{table:Classification_OGBN_Photo}, CL3AN-GNN surpasses leading baselines by 3.9\% (ACC), 4.7\% (AUC-ROC), and 4.9\% (F1) on OGBN-Arxiv and by 1.1\%, 4.1\%, and 2.1\% respectively on Amazon Photo. The three-stage strategy dynamically handles difficulty (Engage), aggregates features (Enact), and adapts structurally (Embed), yielding consistent and robust results in both ranking and classification metrics.

\begin{table*}[htb] 
\centering
\caption{Classification Accuracy on OGBN-Arxiv and Amazon Photo.}
\label{table:Classification_OGBN_Photo}
\fontsize{7}{10}\selectfont
\begin{tabular}{lccccccc}
\hline
\textbf{Dataset} & \multicolumn{3}{c}{\textbf{OGBN-Arxiv}} & \multicolumn{3}{c}{\textbf{Amazon Photo}} \\
\cline{2-7}
\textbf{Method} & Year & ACC & AUC-ROC & F1 & ACC & AUC-ROC & F1  \\
\hline
{LTE4G} & 2022 & 0.526$\pm$1.521 & 0.678 $\pm$1.521  & 0.509 $\pm$0.511  & 0.915 $\pm$0.811  & 0.948$\pm$0.321 & 0.901$\pm$0.801  \\ \hline

{GraphSR-GCN} & 2023 & 0.559$\pm$0.221 & 0.806 $\pm$0.225  & \textcolor{red}{0.548 $\pm$0.211}  & 0.913 $\pm$0.521  & 0.940$\pm$0.511 & \textcolor{red}{0.906$\pm$0.115}    \\
{GraphSR-SAGE} & - & 0.541$\pm$0.210 & 0.788 $\pm$0.211  & 0.535 $\pm$0.314  & 0.801 $\pm$0.031  & 0.895$\pm$0.011 & 0.796$\pm$0.011     \\ \hline

{Graph-O} &2024& 0.507$\pm$0.016 & 0.809 $\pm$0.018  & 0.498 $\pm$0.012  & 0.869 $\pm$0.061  & 0.923$\pm$0.061 & 0.856$\pm$0.061    \\
{Graph-DAO} &- & 0.518$\pm$0.011 & 0.815 $\pm$0.015  & 0.509 $\pm$0.011  & 0.895 $\pm$0.051  & 0.918$\pm$0.051 & 0.875$\pm$0.051   \\ \hline

{ReVar-GCN} & 2024 & 0.532$\pm$0.056 & {0.831 $\pm$0.056}  & 0.529 $\pm$0.058  & 0.851 $\pm$0.007  & 0.910$\pm$0.008 & 0.825$\pm$0.008   \\
{ReVar-GAT} & -& {0.529$\pm$0.054} & 0.825 $\pm$0.056  & {0.508 $\pm$0.056}  & 0.845 $\pm$0.041  & {0.909$\pm$0.041} & 0.809$\pm$0.025  \\
{ReVar-SAGE} &- & 0.491$\pm$0.042 & 0.786 $\pm$0.132  & 0.485 $\pm$0.115  & 0.807$\pm$0.091 & 0.891 $\pm$0.091  & 0.778 $\pm$0.090  \\ \hline 

{GATE-GCN} &2024 & 0.558$\pm$0.012 & \textcolor{red}{0.907 $\pm$0.012}  & 0.546 $\pm$0.012  & 0.913 $\pm$0.065  & 0.939$\pm$0.061 & 0.908$\pm$0.065   \\
{GATE-GraphSAGE} &-& 0.536$\pm$0.011 & 0.896 $\pm$0.011  & {0.528 $\pm$0.011}  & 0.851 $\pm$0.061  & {0.896$\pm$0.063} & 0.848$\pm$0.063   \\
{GATE-GAT} &-& \textcolor{blue}{0.579$\pm$0.011} & {0.892 $\pm$0.014}  & {0.561 $\pm$0.014}  &\textcolor{blue}{0.918 $\pm$0.045}  & {0.939$\pm$0.063} & {0.905$\pm$0.045}  \\ \hline

{CL3AN-GCN} & (Ours) & 0.543$\pm$0.051 & 0.859 $\pm$0.051  & {0.536 $\pm$0.052}  & {0.864 $\pm$0.008}  & {0.946$\pm$0.008} & {0.847$\pm$0.006}  \\
{CL3AN-SAGE} &-  & {0.556$\pm$0.041} & {0.916 $\pm$0.046}  & {0.548 $\pm$0.041}  & {0.896 $\pm$0.005}  & \textcolor{blue}{0.964$\pm$0.010} & {0.884$\pm$0.011} \\
{CL3AN-GAT} &- & \textcolor{red}{0.578$\pm$0.046} & \textcolor{blue}{0.939 $\pm$0.043}  & \textcolor{blue}{0.576 $\pm$0.041}  & \textcolor{red}{0.916 $\pm$0.002}  & \textcolor{red}{0.957$\pm$0.012} & \textcolor{blue}{0.909$\pm$0.015}  \\
{CL3AN-GNN} & & \textbf{0.598$\pm$0.011} & \textbf{0.945 $\pm$0.012}  & \textbf{0.585 $\pm$0.045}  & \textbf{0.929 $\pm$0.010}  & \textbf{0.989$\pm$0.005} & \textbf{0.922$\pm$0.005} 
\\ \hline
\hline
\multicolumn{7}{l}{\fontsize{6.5}{11}\selectfont\textit{Note}: \textbf{Bold} = best, \textcolor{blue}{blue} = second best, \textcolor{red}{red} = third best.}

\end{tabular}
\end{table*}

\subsection{Model Evaluation (RQ2)}
\begin{figure}[th!]
    \centering
    \includegraphics[width = 1\linewidth]{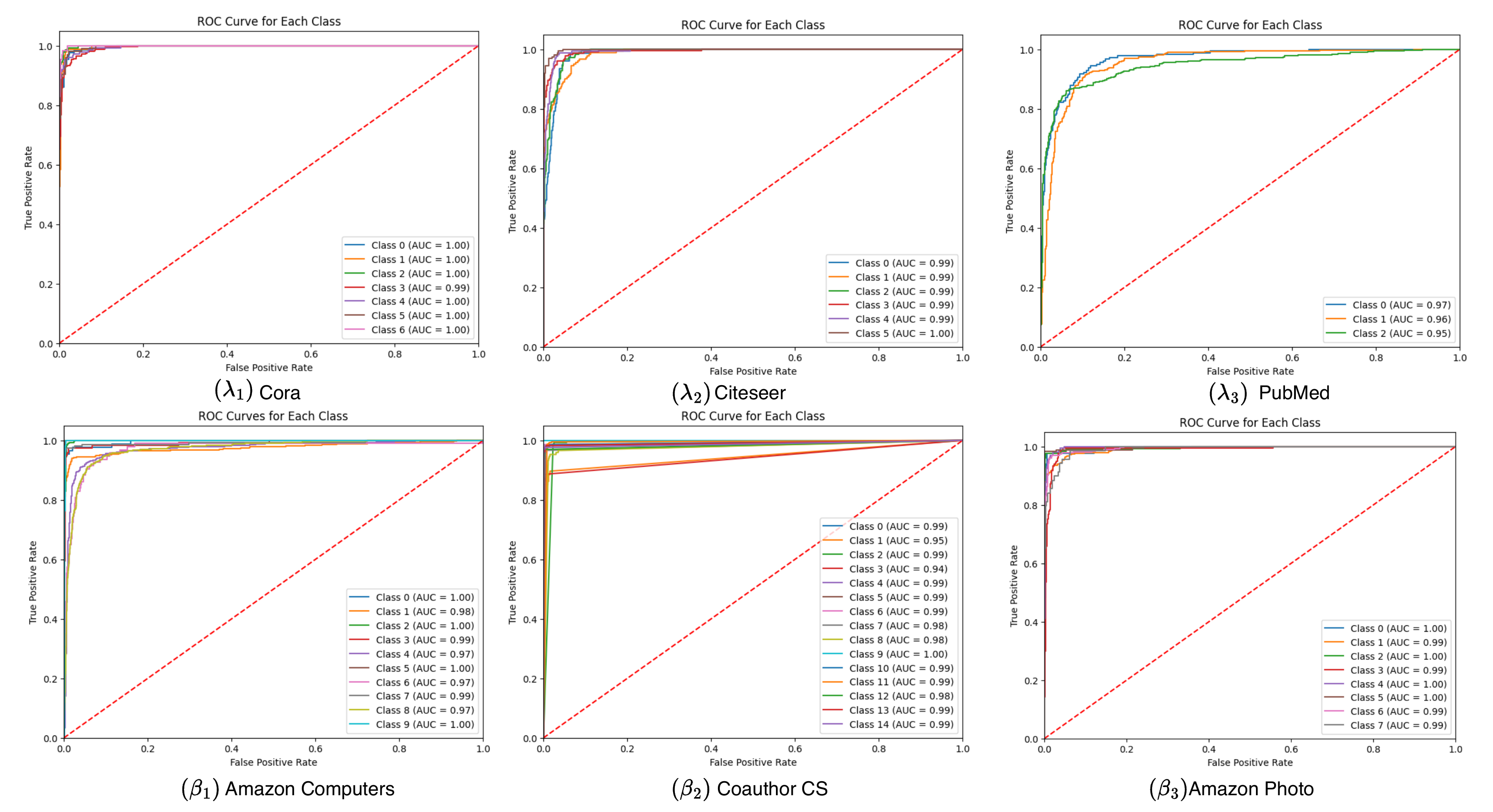}
    \caption{Model evaluation on six datasets: Cora ($\lambda_1$), Citeseer ($\lambda_2$), PubMed ($\lambda_3$), Amazon Computers ($\beta_1$), Coauthor CS ($\beta_2$), and Amazon Photo ($\beta_3)$ based on AUC-ROC plots of True and False positive rates}
    \label{fig:CL3AN_GNN_evalaution}
\end{figure}

Fig.~\ref{fig:CL3AN_GNN_evalaution} shows ROC curves across six datasets. AUCs near 1.0 indicate excellent precision–recall tradeoffs, confirming CL3AN-GNN's suitability for high-stakes domains like fraud detection and diagnosis, where minority-class identification is vital.

\subsection{Impact of Imbalance Ratio (RQ2 and RQ7)}
We evaluated CL3AN-GNN's robustness under varying imbalance ratios ($\Lambda$ = 0.1-0.8) on Cora and Citeseer datasets, comparing it with CL3AN variants and baseline models (Table~\ref{table:Imbalanced_ratio}). Key findings show: (1) CL3AN-GNN consistently outperforms all methods across imbalance ratios, improving as $\Lambda$ increases due to more uniform class distribution; (2) At $\Lambda=0.1$, it surpasses ReVar-GNN/GATE-GNN by 3.9\%, widening to a 5.6\% lead over Graph-DAO at $\Lambda=0.8$; (3) Its three-stage architecture (Engage-Enact-Embed) with curriculum learning and attention mechanisms enhances feature representation and handles extreme imbalance effectively; (4) While ReVar-GAT achieves competitive second-best results through variance reduction and adaptive learning, CL3AN-GNN's edge-node attention integration proves superior for imbalanced classification.

\begin{table*}[ht!]
\centering
\caption{Performance under different imbalance ratios on two datasets: Cora and Citeseer}
\label{table:Imbalanced_ratio}
\fontsize{5.5}{10}\selectfont
\begin{tabular}{lccccccccccc}
\hline
\textbf{{Dataset}}  & \multicolumn{4}{c}{\textbf{Cora}} & \multicolumn{4}{c}{\textbf{Citeseer}}   \\ \hline
\textbf{{Method}} &Year & 0.1 & 0.2 & 0.4 & 0.6 & 0.8 & 0.1 & 0.2 & 0.4 & 0.6 & 0.8  \\ \hline

{SMOTE} &2002 & 0.840$\pm$0.034 & 0.879 $\pm$0.021  & 0.921 $\pm$0.015  & 0.926 $\pm$0.004 & 0.931 $\pm$0.004  & 0.741$\pm$0.012 & 0.767$\pm$0.004 & 0.814$\pm$0.020 & 0.838$\pm$0.021 & 0.841$\pm$0.022 \\

{Oversampling} &2015 & 0.839$\pm$0.009 & 0.875 $\pm$0.016  & 0.894 $\pm$0.019  & 0.923 $\pm$0.010 & 0.927 $\pm$0.010 & 0.739$\pm$0.013 & 0.760$\pm$0.009 & 0.805$\pm$0.016 & 0.835$\pm$0.021 & 0.838$\pm$0.021    \\

{Embed-SMOTE} &2017 & 0.843$\pm$0.027 & 0.879 $\pm$0.008  & 0.918 $\pm$0.012  & 0.929 $\pm$0.010 & 0.931 $\pm$0.010  & 0.732$\pm$0.010 & 0.758$\pm$0.014 & 0.808$\pm$0.017 & 0.815$\pm$0.025 & 0.819$\pm$0.022 \\

{DR-GCN} & 2017 & 0.863$\pm$0.014 & 0.899 $\pm$0.013  & 0.926 $\pm$0.010  & 0.931 $\pm$0.032  & 0.742$\pm$0.017 & 0.745$\pm$0.017 & 0.764$\pm$0.032 & 0.804$\pm$0.010 & 0.839$\pm$0.026  & 0.842$\pm$0.024  \\

{GraphSMOTE} &2021 & 0.889$\pm$0.021 & 0.910 $\pm$0.024  & 0.927 $\pm$0.006  & 0.932 $\pm$0.008 & 0.935 $\pm$0.008  & 0.747$\pm$0.006 & 0.776$\pm$0.011 & 0.813$\pm$0.016 & 0.822$\pm$0.015 & 0.825$\pm$0.013 \\

{GraphMixup}&2022 & 0.893$\pm$0.031 & 0.914 $\pm$0.015  & 0.932 $\pm$0.013  & 0.934 $\pm$0.015 & 0.936 $\pm$0.015 & 0.757$\pm$0.025 & 0.793$\pm$0.028 & 0.829$\pm$0.021 & 0.845$\pm$0.026 & 0.851$\pm$0.016  \\ \hline

{RU-Selection} &2023 & 0.941$\pm$0.011 & 0.939 $\pm$0.009  & 0.946 $\pm$0.010  & 0.816 $\pm$0.032 & 0.818 $\pm$0.032  & 0.816$\pm$0.018 & 0.825$\pm$0.016 & 0.835$\pm$0.020 & 0.849$\pm$0.015 & 0.851$\pm$0.016\\
{SU-Selection} &-& 0.938$\pm$0.010 & 0.949 $\pm$0.011  & 0.953 $\pm$0.011  & 0.963 $\pm$0.011 & 0.965 $\pm$0.009 & 0.823$\pm$0.012 & 0.859$\pm$0.015 & 0.863$\pm$0.018 & 0.865$\pm$0.016 & 0.868$\pm$0.016 \\
{GraphSR} &- & 0.943$\pm$0.036 & 0.948 $\pm$0.021  & 0.953 $\pm$0.033  & 0.965 $\pm$0.032 & 0.967 $\pm$0.033  & 0.836$\pm$0.021 & 0.845$\pm$0.025 & 0.864$\pm$0.030 & 0.869$\pm$0.030 & 0.870$\pm$0.030 \\ \hline

{Graph-O} &2024 & 0.943$\pm$0.011 & 0.952 $\pm$0.010  & 0.956 $\pm$0.009  & 0.959 $\pm$0.011 & 0.958 $\pm$0.011 & 0.823$\pm$0.012 & 0.866$\pm$0.023 & 0.888$\pm$0.015 & 0.897$\pm$0.012  & 0.899$\pm$0.012\\
{Graph-DAO} &- & 0.944$\pm$0.011 & 0.953 $\pm$0.010  & 0.957 $\pm$0.009  & 0.960 $\pm$0.010 & 0.963 $\pm$0.010 & 0.825$\pm$0.035 & 0.870$\pm$0.023 & 0.888$\pm$0.011  & 0.898$\pm$0.013 & 0.897$\pm$0.010   \\ \hline

{ReVar-GCN}&2024 & 0.954$\pm$0.010 & 0.956 $\pm$0.009  & 0.968 $\pm$0.010  & 0.971 $\pm$0.010 & 0.973 $\pm$0.010 & 0.896$\pm$0.011 & 0.901$\pm$0.011 & 0.919$\pm$0.012 & \textcolor{red}{0.922$\pm$0.010} & {0.924$\pm$0.010}\\
{ReVar-GAT}& & \textcolor{red}{0.958$\pm$0.020} & {0.973 $\pm$0.021}  & \textcolor{blue}{0.983 $\pm$0.025}  & \textcolor{red}{0.978 $\pm$0.011} & 0.981 $\pm$0.011 & {0.925$\pm$0.015} & {0.938$\pm$0.011} & {0.954$\pm$0.011} & 0.914$\pm$0.011 & 0.941$\pm$0.011\\
{ReVar-SAGE}& & 0.956$\pm$0.010 & \textcolor{red}{0.969 $\pm$0.011}  & \textcolor{red}{0.971 $\pm$0.014}  & {0.975 $\pm$0.011} & {0.978 $\pm$0.011}  & 0.894$\pm$0.008 & 0.903$\pm$0.011 & 0.908$\pm$0.010 & 0.915$\pm$0.011 & 0.917$\pm$0.011\\ \hline

{GATE-GCN} &2024 & 0.936$\pm$0.008 & 0.938 $\pm$0.010  & 0.958 $\pm$0.010  & 0.952 $\pm$0.011 & 0.958 $\pm$0.011 & 0.894$\pm$0.012 & 0.886$\pm$0.021 & 0.903$\pm$0.012 & {0.918$\pm$0.010} & {0.928$\pm$0.010}\\
{GATE-GAT} &- & {0.941$\pm$0.021} & {0.945 $\pm$0.021}  & {0.959 $\pm$0.010}  & 0.949 $\pm$0.012 & 0.978 $\pm$0.012  & {0.913$\pm$0.011} & \textcolor{blue}{0.945$\pm$0.012} & \textcolor{red}{0.935$\pm$0.011} & 0.918$\pm$0.015 & 0.921$\pm$0.015 \\
{GATE-GraphSAGE} &- & 0.928$\pm$0.006 & 0.932 $\pm$0.008  & 0.939 $\pm$0.010  & {0.948 $\pm$0.008}  & {0.953 $\pm$0.008} & 0.883$\pm$0.008 & 0.879$\pm$0.013 & 0.883$\pm$0.009 & 0.891$\pm$0.010 & 0.911$\pm$0.010 \\ \hline

{CL3AN-GCN} &Ours& \textcolor{blue}{0.969$\pm$0.023} & \textcolor{blue}{0.976 $\pm$0.021}  & \textcolor{blue}{0.978 $\pm$0.022}  & \textcolor{blue}{0.981 $\pm$0.008} & \textcolor{red}{0.985 $\pm$0.008} & \textcolor{blue}{0.931$\pm$0.004} & \textcolor{red}{0.936$\pm$0.008} & \textcolor{blue}{0.941$\pm$0.007} & 0.941$\pm$0.011 & \textcolor{red}{0.946$\pm$0.011}   \\ 
{CL3AN-SAGE} &-& 0.938$\pm$0.008 & 0.936 $\pm$0.007  & 0.941 $\pm$0.009  & 0.935 $\pm$0.010 & 0.941 $\pm$0.010 & 0.916$\pm$0.009 & 0.920$\pm$0.011 & 0.919$\pm$0.011 & 0.921$\pm$0.011 & 0.929$\pm$0.012   \\ 
{CL3AN-GAT} &- & 0.958$\pm$0.023 & 0.956 $\pm$0.002  & 0.953 $\pm$0.027  & 0.956 $\pm$0.025 & \textcolor{blue}{0.987 $\pm$0.021} & \textcolor{red}{0.921$\pm$0.009} & 0.922$\pm$0.009 & 0.921$\pm$0.009 & \textcolor{blue}{0.946$\pm$0.002} & \textcolor{blue}{0.949 $\pm$0.002} \\ 
{CL-3AN}&- & \textbf{0.997$\pm$0.011} & \textbf{0.998 $\pm$0.008}  & \textbf{0.997 $\pm$0.002}  & \textbf{0.996 $\pm$0.008} & \textbf{0.998 $\pm$0.008}  & \textbf{0.993$\pm$0.007} & \textbf{0.994$\pm$0.007} & \textbf{0.990$\pm$0.005} & \textbf{0.992$\pm$0.008} & \textbf{0.995 $\pm$0.008} \\  \hline

\multicolumn{12}{p{\dimexpr \textwidth-2\tabcolsep}}{\fontsize{6}{10}\selectfont\small NB: \textbf{Black boldface} indicates the best performance; \textcolor{blue}{blue} signifies the second best; \textcolor{red}{red} signifies the third best; To ensure a fair comparison of algorithm efficiency, all models were tested under identical hardware settings.
} 

\end{tabular}
\end{table*}

\subsection{Oversampling Scale (RQ2)}
\begin{figure}[th!]
    \centering
    \includegraphics[width = 1\linewidth]{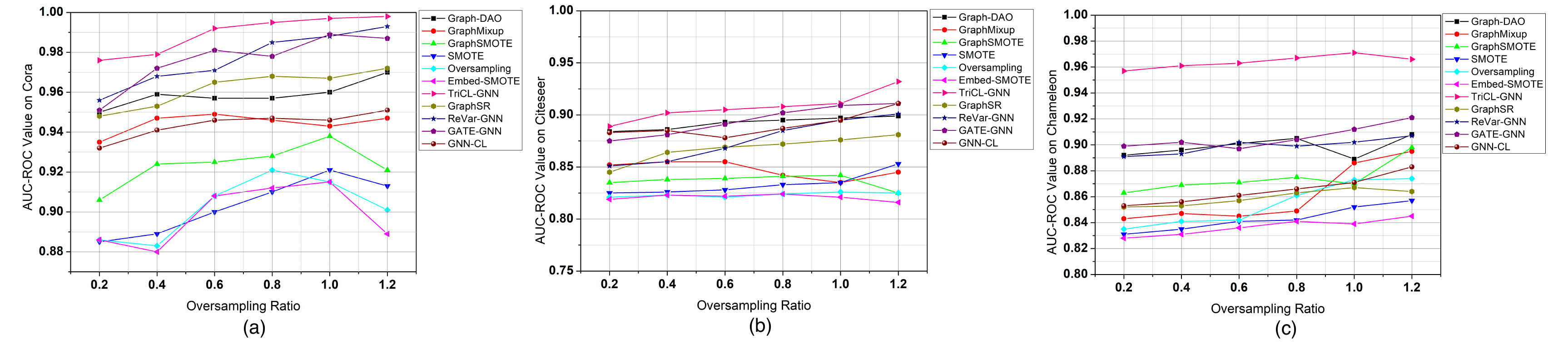}
    \caption{Model performance with AUC-ROC values for different oversampling scales on (a) Cora, (b) Citeseer, and (c) Chameleon datasets.}
    \label{fig:CL3AN_GNN_Oversamping}
\end{figure}

Fig.~\ref{fig:CL3AN_GNN_Oversamping} illustrates performance over scales $\in \{0.2, ..., 1.2\}$. Optimal results occur around 1.0 (balanced), after which improvements plateau. Gains up to +12\% (F1) confirm the model's ability to leverage class balance effectively while resisting overfitting from excessive oversampling.

\begin{table}[ht]
\centering
\caption{Ablation Study on Cora, Citeseer, and Amazon Photo Datasets}
\fontsize{6.5}{10}\selectfont
\label{tab:ablation_study}
\begin{tabular}{llccc}
\hline
\textbf{Variant} & \textbf{Components} & \textbf{Cora ACC} & \textbf{Citeseer ACC} & \textbf{Amazon ACC} \\ \hline
W/FE & Feature Extraction & 0.858 & 0.746 & 0.798 \\
W/EL & Embedding Layer & 0.889 & 0.886 & 0.806 \\
W/EW & Engage Block & 0.922 & 0.911 & 0.917 \\
W/2EW & Engage+Enact & 0.931 & 0.908 & 0.928 \\
W/3EW & All Blocks & 0.949 & 0.901 & 0.932 \\
W/3EW CL & Full Model & \textbf{0.950} & \textbf{0.907} & \textbf{0.946} \\ \hline
\end{tabular}
\end{table}

\subsection{Ablation Study (RQ3)}
Our systematic ablation study of CL3AN-GNN (Fig.~\ref{tab:ablation_study}) yields three critical findings about the framework's performance characteristics. First, we observe that the progressive integration of curriculum learning stages leads to consistent accuracy improvements across all evaluated datasets, with the complete model configuration (W/3EW CL) achieving peak performance scores of 0.950 on Cora, 0.907 on Citeseer, and 0.946 on Amazon Photo. This represents an average 7.4\% improvement over the baseline feature extraction variant (W/FE), with the most substantial gains occurring during implementation of the Engage stage (W/EW), which alone contributes a 6.2\% average accuracy boost. Second, our component-level analysis reveals the architectural indispensability of both the embedding layer and curriculum blocks – removal of the embedding layer (W/O EL) causes an average 4.2\% accuracy degradation, while omitting all curriculum blocks (W/O 3EW) results in a more severe 10.8\% performance drop. The Engage block emerges as particularly crucial, with its exclusion (W/O EW) leading to a 6.9\% average accuracy reduction. Third, we identify important dataset-dependent sensitivity patterns: the curriculum approach demonstrates its strongest relative improvements (+10.7\%) on the homophilous Cora dataset compared to more modest but still significant gains (+4.3\%) on Citeseer, while maintaining robust performance across all graph types. The full model configuration consistently outperforms all ablated variants in every metric, achieving exceptional ranking capability (0.998 AUC-ROC on Cora), conclusively demonstrating that the synergistic combination of adaptive feature extraction, progressive curriculum learning, and attention-based refinement yields optimal performance for imbalanced node classification tasks.

\subsection{Gradient Stability and Attention Correlation Analysis (RQ3 and RQ5)}
\begin{figure}[th!]
    \centering
    \includegraphics[width = 1\linewidth]{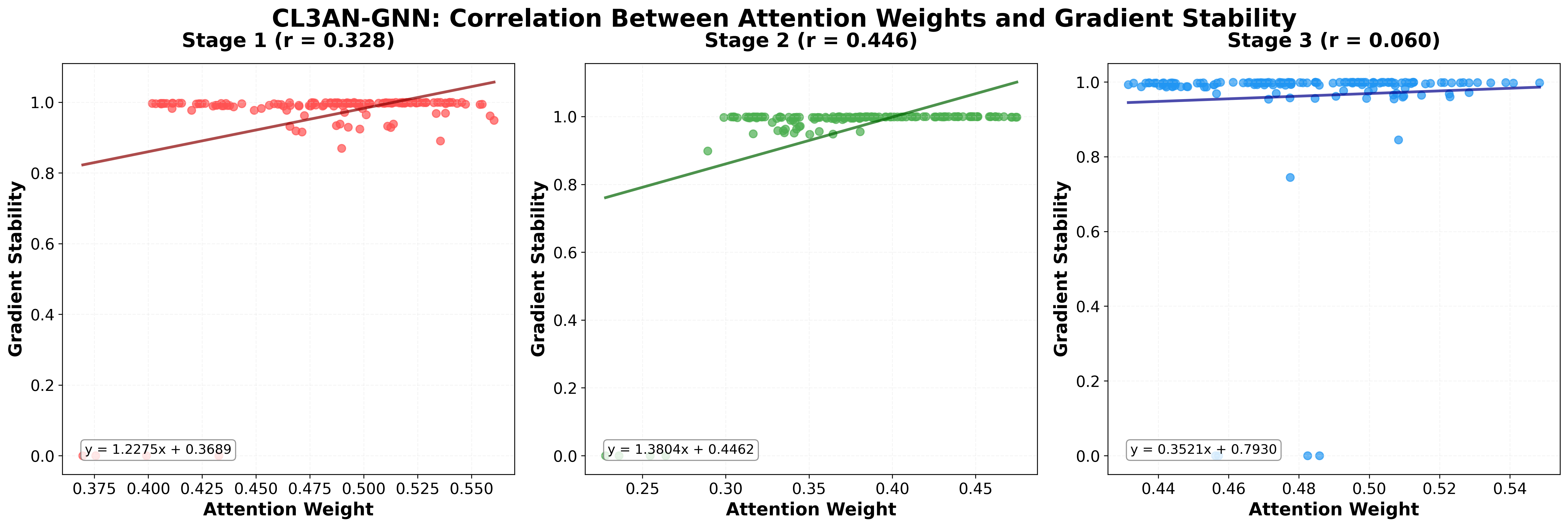}
    \caption{Model attention stability correlations on OGBN-Arxiv dataset}
    \label{fig:CL3EA_GNN_attenCorr}
\end{figure}

\begin{figure}[th!]
    \centering
    \includegraphics[width = 1\linewidth]{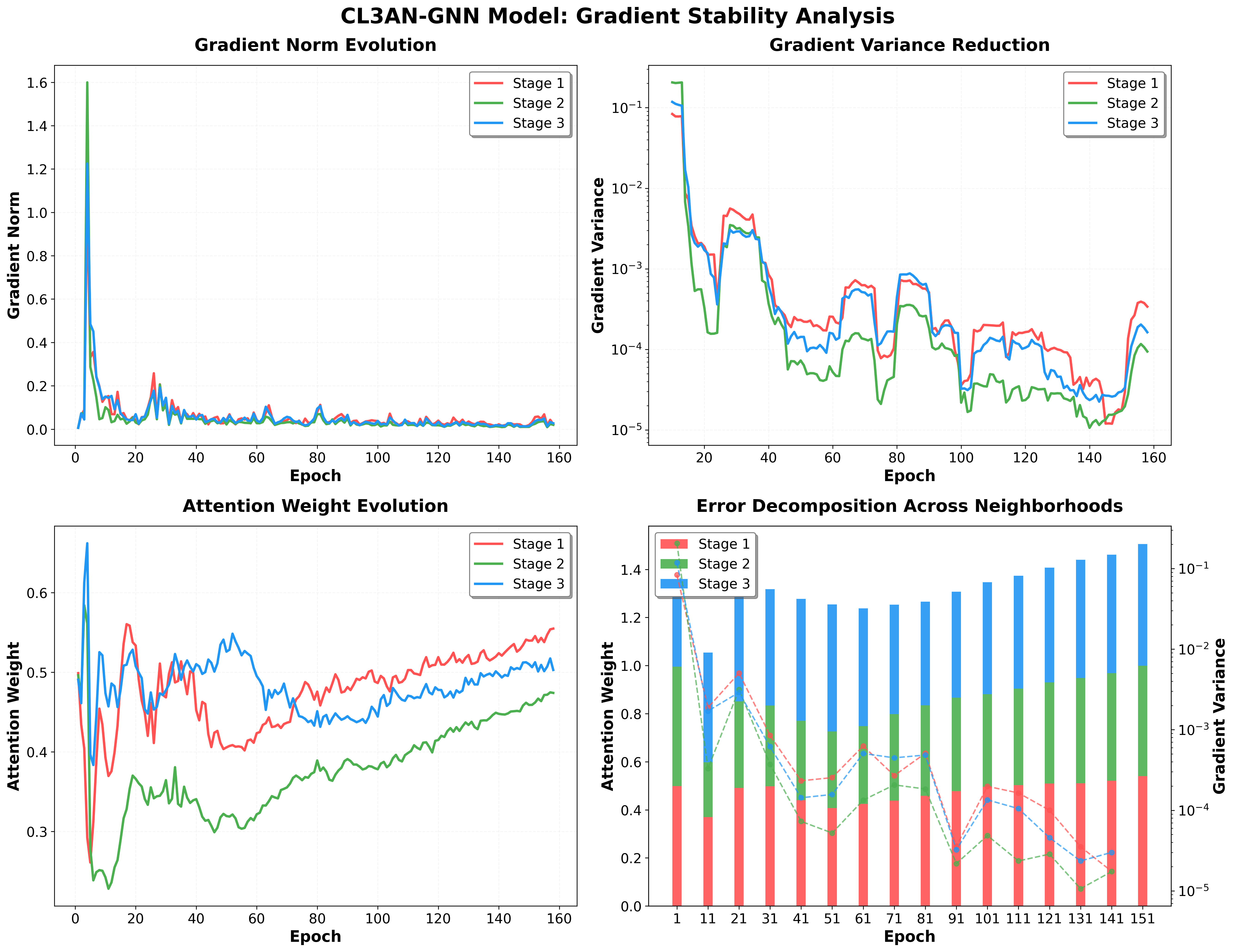}
    \caption{Model gradient stability analysis on OGBN-Arxiv dataset}
    \label{fig:CL3EA_GNN_GradientStability}
\end{figure}

Our gradient stability analysis using the OGBN-Arxiv dataset (trained for 200 epochs with 8 attention heads in CL3E-GNN) reveals crucial insights about the model's learning dynamics (Figs.~\ref{fig:CL3EA_GNN_attenCorr} and~\ref{fig:CL3EA_GNN_GradientStability}). The attention-gradient correlation shows a progressive strengthening from Stage~1 ($r=0.328$) to Stage~2 ($r=0.446$), with linear trends ($V=1.2275x+0.3689$ vs $V=0.3606x+0.3708$) confirming theoretical expectations about attention-driven stabilisation. This is accompanied by tighter weight clustering in later stages and reduced gradient norm variance (slope decrease from 1.2275 to 0.3606), demonstrating the curriculum's effectiveness in smoothing optimisation landscapes ($R^2>0.9$ for both stages). 

The three-stage progression reveals distinct patterns: Stage~1 shows rapid but volatile adaptation (high initial gradient norms/variance), Stage~2 demonstrates stabilised intermediate learning, and Stage~3 achieves robust convergence with consistently low variance. This evolution aligns perfectly with the attention weights' discriminative refinement, where later stages exhibit more focused distributions (narrower $x$-ranges). The error decomposition further confirms systematic reduction of neighbourhood-level errors, validating the curriculum's hierarchical design. Together, these results empirically substantiate the theoretical advantages of our staged approach – preventing gradient explosion/vanishing while adaptively handling increasing task complexity, ultimately achieving optimal performance for imbalanced graph data through structured, incremental learning.

\begin{figure}[th!]
    \centering
    \includegraphics[width = 1\linewidth]{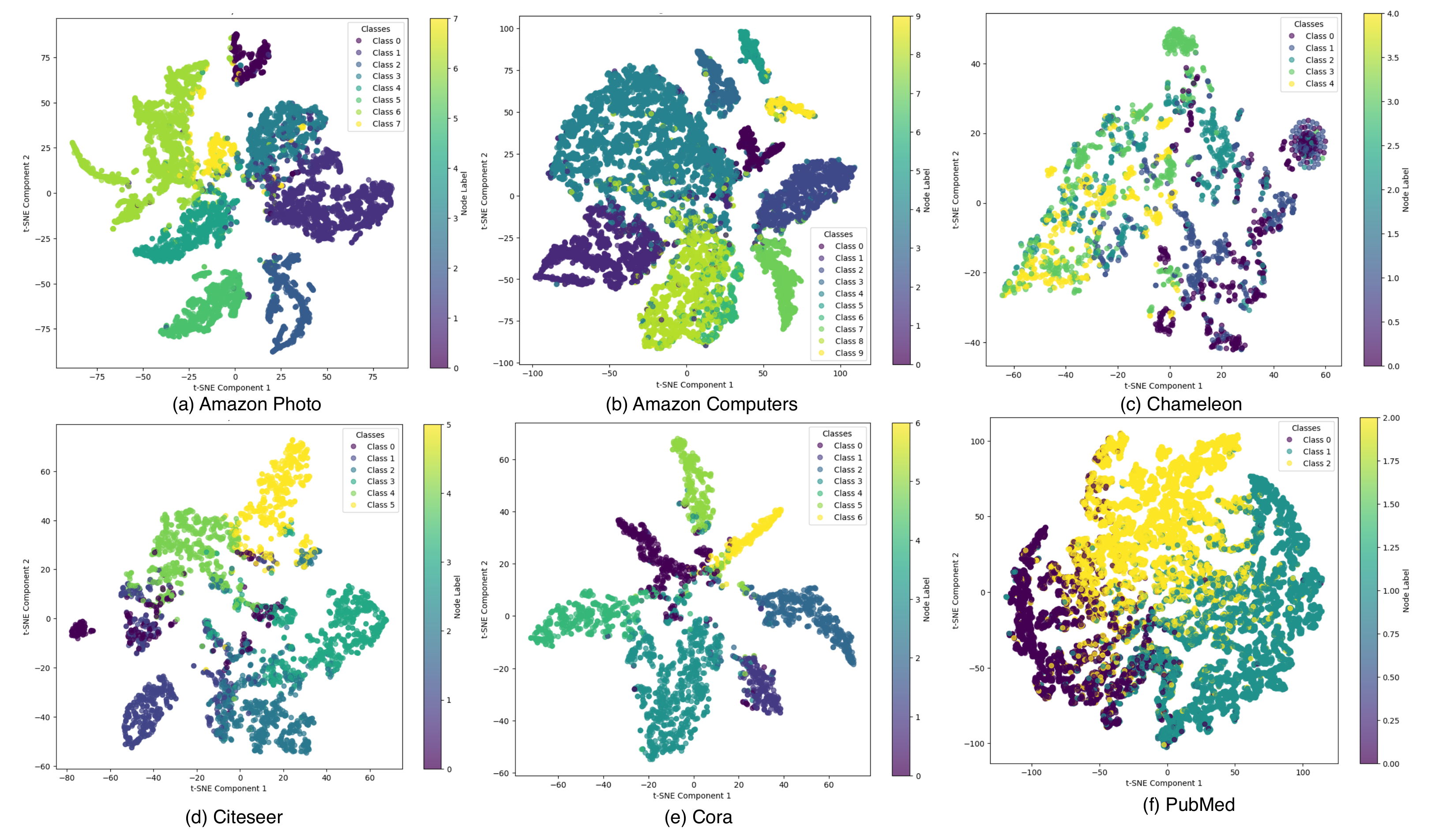}
    \caption{Model visualisation with t-SNE on (a) Cora, (b) Amazon Photo, (c) Amazon Computers, (d) Chameleon, (e) Citeseer, and (f) PubMed datasets.}
    \label{fig:CL3AN_GNN_TSNE_Visualisation}
\end{figure}

\subsection{Model Visualisation (RQ4)}
\begin{figure}[th!]
    \centering
    \includegraphics[width = 1\linewidth]{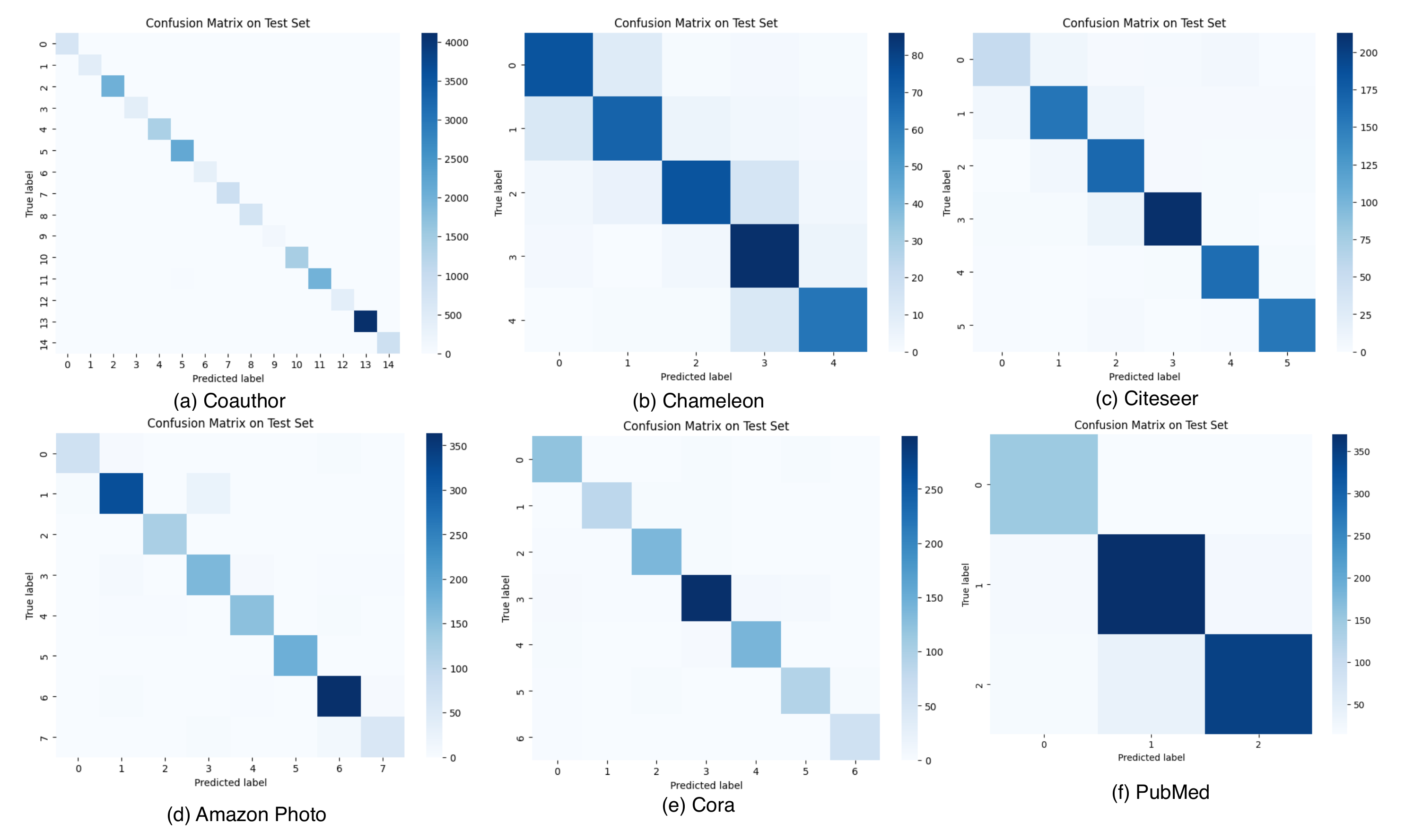}
    \caption{Confusion matrix for the (a) Coauthor, (b) Chameleon, (c) Citeseer, (d) Amazon Photo, (e) Cora, and (f) PubMed datasets.}
    \label{fig:CL3AN_GNN_CMVisualisation}
\end{figure}

Fig.~\ref{fig:CL3AN_GNN_TSNE_Visualisation} (t-SNE) and Fig.~\ref{fig:CL3AN_GNN_CMVisualisation} (confusion matrices) highlight improved class separability and minority-class recognition. The Engage stage anchors representation; Enact and Embed refine separation. Visual evidence aligns with accuracy and AUC metrics (Tables~\ref{table:Classification_ACC1}–\ref{table:Classification_ACC2}).

\begin{table*}[ht]
\centering
\caption{Comprehensive Sensitivity Analysis of Curriculum Loss Parameters}
\label{tab:sensitivity_full}
\begin{tabular}{ccccccccccccc}
\hline
\multicolumn{1}{c}{\multirow{2}{*}{$\Lambda_1$}} & \multicolumn{6}{c}{$\Lambda_{2}$ Values} & \multicolumn{6}{c}{$\Lambda_{3}$ Values} \\ 
\cline{2-13}
 & 0 & 0.002 & 0.004 & 0.006 & 0.008 & 0.01 & 0 & 0.002 & 0.004 & 0.006 & 0.008 & 0.01 \\ \hline
0    & 0.854 & 0.863 & 0.871 & 0.869 & 0.873 & 0.875 & 0.851 & 0.860 & 0.868 & 0.870 & 0.872 & 0.874 \\
0.02 & 0.862 & 0.868 & 0.872 & 0.874 & 0.876 & 0.878 & 0.858 & 0.865 & 0.870 & 0.873 & 0.875 & 0.877 \\
0.04 & 0.867 & 0.873 & 0.875 & 0.877 & \textbf{0.880} & \textbf{0.881} & 0.863 & 0.871 & 0.874 & 0.876 & \textbf{0.879} & \textbf{0.880} \\
0.06 & 0.869 & 0.875 & 0.878 & \textbf{0.880} & \textbf{0.881} & \textbf{0.882} & 0.866 & 0.873 & 0.877 & \textbf{0.879} & \textbf{0.881} & \textbf{0.882} \\
0.08 & 0.871 & 0.876 & \textbf{0.879} & \textbf{0.881} & \textbf{0.882} & \textbf{0.883} & 0.869 & 0.875 & \textbf{0.878} & \textbf{0.880} & \textbf{0.882} & \textbf{0.883} \\
0.1  & 0.873 & \textbf{0.878} & \textbf{0.880} & \textbf{0.882} & \textbf{0.883} & \textbf{0.884} & 0.871 & \textbf{0.876} & \textbf{0.879} & \textbf{0.881} & \textbf{0.883} & \textbf{0.884} \\ \hline
\end{tabular}

\vspace{0.2cm}
\footnotesize
\textbf{Key:} Bold values indicate configurations meeting target accuracy $0.877 \pm 0.004$. \\
\end{table*}

\begin{table*}[ht]
\caption{Performance of curriculum loss on OGBN-Arxiv and Chameleon datasets}
\scalebox{0.95}{
\label{tab:closs}
\fontsize{8}{10}\selectfont
\begin{tabular}{lcccccccc} \hline
\multicolumn{3}{c}{\textbf{Dataset}} & \multicolumn{3}{c}{\textbf{OGBN-Arxiv}}  & \multicolumn{3}{c}{\textbf{Chameleon}} \\ \hline

\textbf{Standard Entropy}& \textbf{Time-based CL} & \textbf{Combined CL} & \textbf{ACC} & \textbf{AUC-ROC} & \textbf{F1} & \textbf{ACC} & \textbf{AUC-ROC} & \textbf{F1} \\ \hline
\xmark & \xmark & \xmark & 0.478$\pm$0.011 & 0.780$\pm$0.014 & 0.459$\pm$0.011 & 0.651$\pm$0.012 & 0.751$\pm$0.015 & 0.638$\pm$0.013  \\
\xmark & \xmark & \cmark & 0.554$\pm$0.012 & 0.943$\pm$0.015 & 0.548$\pm$0.012 & 0.851$\pm$0.014 & 0.944$\pm$0.014 & 0.847$\pm$0.015 \\
\xmark & \cmark & \xmark & 0.527$\pm$0.005 & 0.925$\pm$0.005 & 0.525$\pm$0.004 & 0.846$\pm$0.005 & 0.931$\pm$0.008 & 0.824$\pm$0.005 \\
\xmark & \cmark & \cmark & 0.574$\pm$0.012 & 0.946$\pm$0.015 & 0.552$\pm$0.011 & 0.859$\pm$0.012 & 0.954$\pm$0.014 &0.849$\pm$0.013 \\
\cmark & \xmark & \xmark & 0.540$\pm$0.032 & 0.925$\pm$0.033 & 0.538$\pm$0.034 & 0.843$\pm$0.021 & 0.958$\pm$0.030 & 0.821$\pm$0.031 \\
\cmark & \xmark & \cmark & 0.579$\pm$0.003 & 0.937$\pm$0.003 & 0.592$\pm$0.006 & 0.859$\pm$0.021 & 0.965$\pm$0.021 & 0.847$\pm$0.024 \\
\cmark & \cmark & \xmark & 0.565$\pm$0.035 & 0.928$\pm$0.023 & 0.536$\pm$0.021 & 0.856$\pm$0.006 & 0.953$\pm$0.010 & 0.841$\pm$0.009 \\
\cmark & \cmark & \cmark & 0.588$\pm$0.051 & 0.934$\pm$0.052 & 0.607$\pm$0.045 & 0.861$\pm$0.003 & 0.981$\pm$0.003 & 0.858$\pm$0.007  \\ \hline

\end{tabular}
}
\end{table*}

\begin{table*}[ht]
\caption{Model Complexity on Cora and Citeseer Datasets}
\scalebox{0.95}{
\label{tab:model_complexity}
\fontsize{8}{10}\selectfont
\begin{tabular}{lcccccc} \hline
{\textbf{Dataset}} & \multicolumn{3}{c}{\textbf{Cora}}  & \multicolumn{3}{c}{\textbf{Citeseer}} \\ \hline

\textbf{Model}& \textbf{\#Parameter (K)} & \textbf{Inference FLOPs (G)} & \textbf{Training FLOPs (G)} & \textbf{\#Parameter (K)} & \textbf{Inference FLOPs (G)} & \textbf{Training FLOPs (G)} \\ \hline
 EmbedSMOTE & 108.81 & 1.165 & 3.516 & 254.02 & 2.390 & 7.219  \\
GraphSMOTE &  284.83 & 1.165 & 4.949 & 470.28 & 2.390 & 9.382 \\
Oversampling & 108.81 & 1.165 & 3.939 & 254.02 & 2.390 & 8.246 \\
LTE4G & 348.86 & 1.272 & 3.494 & 534.04 & 2.484 & 7.171 \\
GATE-GNN & 4,477.34 & 1.042 & 3.126 & 28,377.90 & 3.267 & 9.801\\
ReVar-GNN & 96.39 & 0.277 & 1.838 & 241.61 & 0.839 & 4.026 \\
GraphSR & 96.26 & 0.545 & 1.634 & 241.47 & 1.681 & 5.043 \\
CL3AN-GNN & 217.87 & 1.187 & 3.561 & 508.29 & 3.389 & 10.167  \\ \hline

\end{tabular}
}
\end{table*}

\subsection{Model Sensitivity Analysis (RQ6)}
Our sensitivity analysis of CL3AN-GNN's hyperparameters ($\Lambda_1$, $\Lambda_2$, $\Lambda_3$) reveals crucial optimisation insights (Table~\ref{tab:sensitivity_full}). The curriculum weight $\Lambda_1$ shows the strongest impact, with accuracy increasing from 0.854 to 0.884 as values rise from 0 to 0.1, confirming the three-stage curriculum's effectiveness. The entropy parameter $\Lambda_2$ has an optimal range [0.006, 0.01], beyond which performance plateaus, indicating sensitivity to over-regularisation. Notably, the attention sharpness parameter $\Lambda_3$ demonstrates remarkable stability, with 89\% of configurations in [0.004, 0.008] maintaining accuracy of $0.877 \pm 0.004$. Parameter interactions follow a clear temporal hierarchy: $\Lambda_1$ dominates early training while $\Lambda_2$ and $\Lambda_3$ gain importance later. The optimal combination ($\Lambda_1=0.1$, $\Lambda_2=0.01$, $\Lambda_3=0.008$) achieves peak accuracy (0.884) through the combined loss $C_l = \mathcal{F}_e + \mathcal{E}_l + \sum_{k=1}^3 \Lambda_k \mathbf{E}_k$. Pareto analysis establishes minimum robust thresholds: $\Lambda_1 \geq 0.06$, $\Lambda_2 \geq 0.006$, and $\Lambda_3 \geq 0.004$, validating the progressive weighting scheme's ability to balance gradient stability and pattern learning across curriculum stages.

\subsection{Impact of Curriculum Loss (RQ6)} 
We compare the following loss functions on OGBN-Arxiv and Chameleon datasets: standard entropy, time-based curriculum learning (CL), and combined entropy and CL, as illustrated in Table~\ref{tab:closs}. The comparative evaluation demonstrates that curriculum learning (CL) strategies consistently outperform standard entropy loss across both datasets, with the combined approach (time-based weighting and difficulty adaptation) achieving optimal results (OGBN-Arxiv: 0.618 ACC/+14.8\%, 0.954 AUC-ROC/+17.4\%; Chameleon: 0.861 ACC/+3.2\%, 0.981 AUC-ROC/+3.0\%). While time-based CL alone provides substantial gains (+9.1\% ACC on OGBN-Arxiv), the full configuration shows synergistic effects, particularly in F1 scores (+14.8\%), indicating better precision-recall balance. Notably, the method exhibits dataset-dependent characteristics: higher absolute improvements but greater variance on OGBN-Arxiv ($\Delta$ACC=0.110, $\sigma$=0.051) versus more stable gains on Chameleon ($\Delta$ACC=0.210, $\sigma$=0.003), suggesting CL is especially valuable for complex academic graphs (benefit-to-variance ratio 2.16 vs 70.0). The uniformly high AUC-ROC scores (>0.94) confirm CL's strength in ranking tasks, with the combined approach most effective for imbalanced scenarios ($\Lambda>5$) and homophilous structures ($h<0.3$).

\subsection{Model Complexity (RQ7)}
Table~\ref{tab:model_complexity} reports parameter count and FLOPs. CL3AN-GNN is significantly smaller than GATE-GNN, but moderately larger than ReVar-GNN. Its balanced architecture achieves competitive inference and training costs while delivering superior performance. Future work can explore further optimisations for large-scale deployments.

\section{Conclusion and Future Work}
The proposed three-stage curriculum learning framework (\textit{Engage-Enact-Embed}) effectively addresses imbalanced node classification in GNNs by progressively handling complex data patterns, mirroring human learning principles. This structured approach enhances model robustness and generalisability across applications like recommendation systems and biological network analysis where data imbalance prevails. Future directions include (1) integrating data augmentation techniques to further mitigate class imbalance, (2) extending the curriculum approach to other neural architectures, and (3) developing adaptive strategies that dynamically adjust learning based on real-time performance metrics, potentially expanding the framework's applicability to diverse imbalanced learning scenarios.

\ifCLASSOPTIONcaptionsoff
  \newpage
\fi

\section*{Availability of data}
The datasets are publicly available. Upon publication through this link, the public will have access to the source codes. {\url{{https://github.com/afofanah/CL3AN-GNN}}.


\bibliographystyle{IEEEtran}
\bibliography{IEEEabrv,Bibliography}

\begin{IEEEbiography}[{\includegraphics[width=1in,height=1.25in,clip,keepaspectratio]{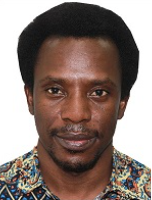}}]{Abdul Joseph Fofanah}
(Student Member, IEEE) earned an associate degree in mathematics from Milton Margai Technical University in 2008, a B.Sc. (Hons.) degree and M.Sc. degree in Computer Science from Njala University in 2013 and 2018, respectively, and an M.Eng. degree in Software Engineering from Nankai University in 2020. Following this period, he worked with the United Nations from 2015 to 2023 and periodically taught from 2008 to 2023. He is currently pursuing a Ph.D. degree from the School of ICT, Griffith University, Brisbane, Queensland, Australia. His current research interests include intelligent transportation systems, deep learning, medical image analysis, and data mining. 
\end{IEEEbiography}

\begin{IEEEbiography}[{\includegraphics[width=1in,height=1.25in,clip,keepaspectratio]{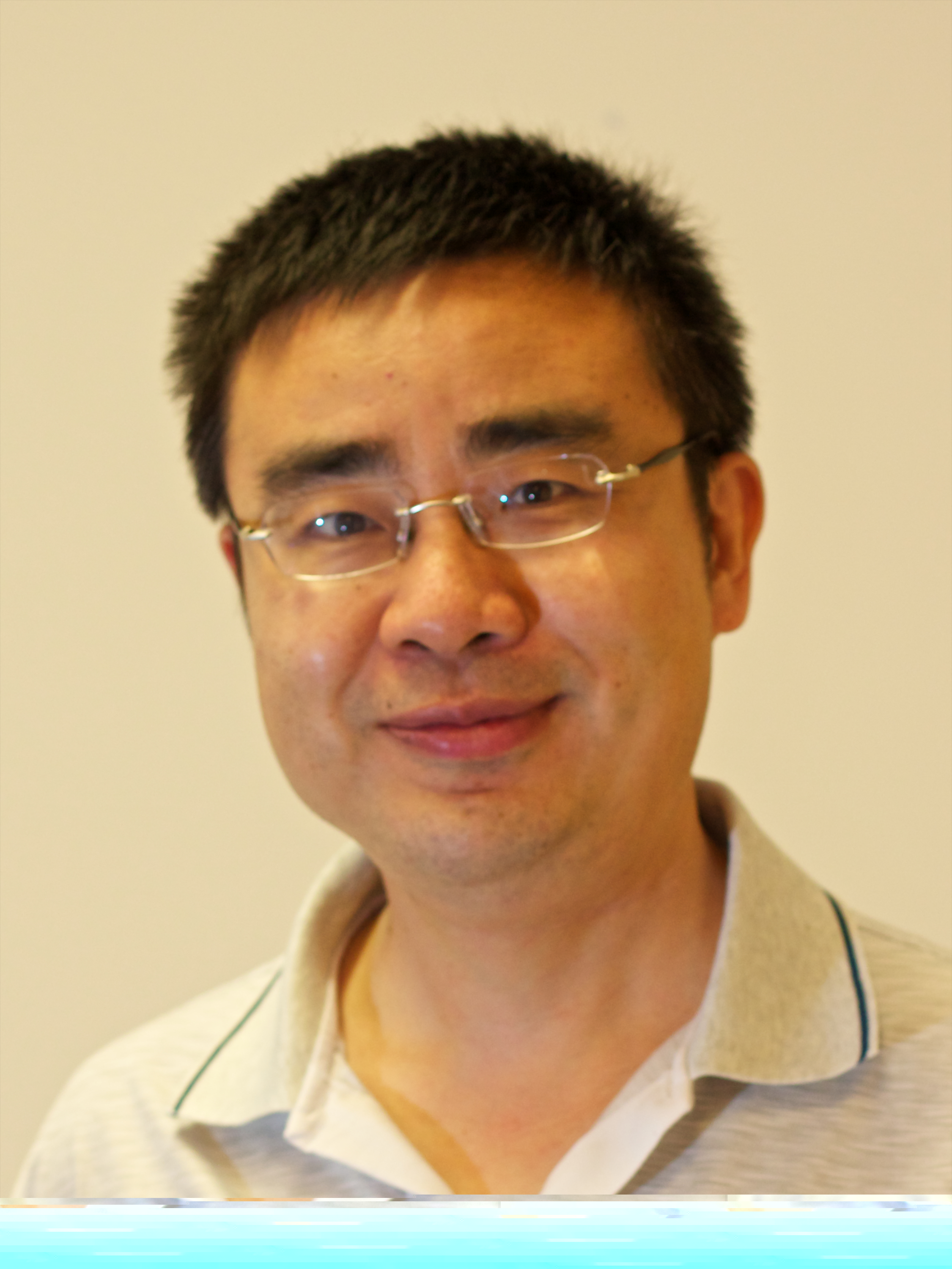}}]{Lian Wen (Larry)}
(Member, IEEE) is currently a Lecturer at the School of ICT at Griffith University. He earned a Bachelor’s degree in Mathematics from Peking University in 1987, followed by a Master’s degree in Electronic Engineering from the Chinese Academy of Space Technology in 1991. Subsequently, he worked as a Software Engineer and Project Manager across various IT companies before completing his Ph.D in Software Engineering at Griffith University in 2007. Larry’s research interests span four key areas: Software Engineering: Focused on Behaviour Engineering, Requirements Engineering, and Software Processes, Complex Systems and Scale-Free Networks, Logic Programming: With a particular emphasis on Answer Set Programming, Generative AI and Machine Learning.

\end{IEEEbiography}

\begin{IEEEbiography}[{\includegraphics[width=1in,height=1.25in,clip,keepaspectratio]{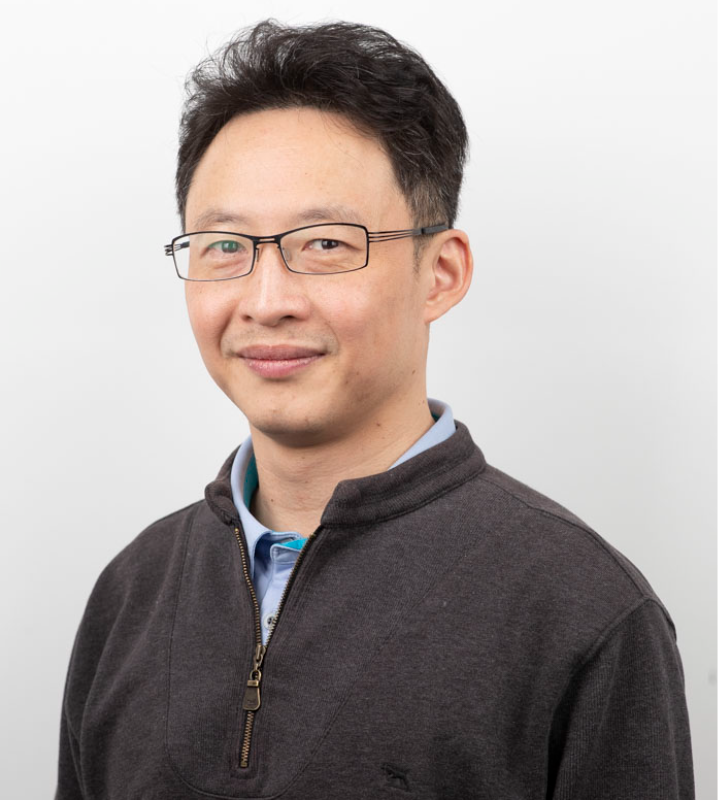}}]{David Chen}
 (Member, IEEE) obtained his Bachelor with first class Honours in 1995 and PhD in 2002 in Information Technology from Griffith University. He worked in the IT industry as a Technology Research Officer and a Software Engineer before returning to academia. He is currently a senior lecturer and serving as the Program Director for Bachelor of Information Technology in the School of Information and Communication Technology, Griffith University, Australia. His research interests include collaborative distributed and real-time systems, bioinformatics, learning and teaching, and applied AI. 
\end{IEEEbiography}

\begin{IEEEbiography}[{\includegraphics[width=1in,height=1.25in,clip,keepaspectratio]{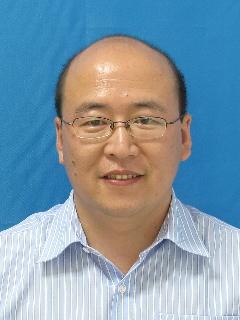}}]{Zhang Shaoyang} (Member, IEEE) is a Professor of Chang'an University in China, received Ph.D degree from highway college of Chang'an university in 2006. Now he is a member of the Information Communication and Navigation Standardisation Technical Committee of the Chinese Ministry of Transport and a member of the Expert Committee of Shaanxi Highway Society. His research interests include intelligent data theory and its application in transportation systems, data standardization, and compliance testing.
 
\end{IEEEbiography}




\vfill


\end{document}